\documentclass[letter,11pt]{article}
 \usepackage{fullpage}

 \usepackage[cmex10]{amsmath}

\usepackage{epsf,psfrag,latexsym,graphicx,bm,xcolor,url,array}
  \usepackage{cases}
\usepackage{verbatim}
\usepackage[mathscr]{eucal}
\usepackage{algorithm, algorithmic}
\usepackage[fleqn,tbtags]{mathtools}
\usepackage{amsbsy,dsfont,amssymb,amsfonts,amsmath}
\usepackage{tikz}
\usetikzlibrary{calc,shapes}
\usepackage{wrapfig}
\usepackage[sort]{natbib}

\newtheorem{theorem}{Theorem}
\newtheorem{corollary}{Corollary}
\newtheorem{lemma}{Lemma}

\newtheorem{proposition}{Proposition}

\newtheorem{fact}{Fact}

\newtheorem{definition}{Definition}

\DeclareMathOperator{\poly}{poly}

  \DeclareMathOperator{\tw}{tw}
  \DeclareMathOperator{\local}{local}
  \DeclareMathOperator{\range}{range}
   \DeclareMathOperator{\Path}{Path}

  \DeclareMathOperator{\nbd}{\mathcal{N}}

 \DeclareMathOperator{\hLambda}{\widehat{\Lambda}}

\DeclareMathOperator{\dist}{dist}
\DeclareMathOperator{\lrank}{LRank}
\DeclareMathOperator{\rank}{Rank}

\def\tha{{\mbox{\tiny th}}}

\DeclareMathOperator{\Diag}{Diag}

\newcommand\indep{\protect\mathpalette{\protect\independenT}{\perp}}
\def\independenT#1#2{\mathrel{\rlap{$#1#2$}\mkern2mu{#1#2}}}

\DeclarePairedDelimiter\norm{\lVert}{\rVert}

 \def\0{{\bf 0}}

\def\viz{{viz.,\ \/}}

\def\st{{s.t.  }}

\def\nn{\nonumber}

\def\qed{\hfill\hbox{${\vcenter{\vbox{
    \hrule height 0.4pt\hbox{\vrule width 0.4pt height 6pt
    \kern5pt\vrule width 0.4pt}\hrule height 0.4pt}}}$}}

\def\tcr{\textcolor{red}}
\def\tcb{\textcolor{blue}}

\def\tccyan{\textcolor{cyan}}
\def\tcv{\textcolor{violet}}
\definecolor{myred}{rgb}{0.3,0.0,0.7}
\definecolor{dkg}{rgb}{0.1,0.7,0.2}
\definecolor{dkb}{rgb}{0.0,0.2,0.8}

\def\tcdkg{\textcolor{dkg}}

 \def\hC{\widehat{C}}

 \def\hG{\widehat{G}}
 
 \def\hI{\widehat{I}}

 \def\hM{\widehat{M}}

 \def\hP{\widehat{P}}

\def\hT{\widehat{T}}
 \def\hU{\widehat{U}}
 \def\hV{\widehat{V}}

\def\bfa{{\mathbf a}}
\def\bfb{{\mathbf b}}

\def\bfe{{\mathbf e}}

\def\bfm{{\mathbf m}}

\def\bfy{{\mathbf y}}
\def\bfz{{\mathbf z}}

\def\bfY{{\mathbf Y}}

\def\lambdabf{\hbox{\boldmath$\lambda$\unboldmath}}

\def\nubf{\hbox{\boldmath$\nu$\unboldmath}}

\def\pibf{\hbox{\boldmath$\pi$\unboldmath}}

\def\Cc{{\cal C}}

\def\Sc{{\cal S}}

\def\Xc{{\cal X}}
\def\Yc{{\cal Y}}

\def\Fbb{{\mathbb F}}

\def\Nbb{{\mathbb N}}

\def\Pbb{{\mathbb P}}

\def\Rbb{{\mathbb R}}

\def\tilC{{\widetilde{C}}}

\newcommand{\bprf}{\begin{myproof}}
\newcommand{\eprf}{\end{myproof}}
\newcommand{\bp}{\begin{psfrags}}
\newcommand{\ep}{\end{psfrags}}
\newcommand{\bl}{\begin{lemma}}
\newcommand{\el}{\end{lemma}}
\newcommand{\bt}{\begin{theorem}}
\newcommand{\et}{\end{theorem}}
\newcommand{\bc}{\begin{center}}
\newcommand{\ec}{\end{center}}
\newcommand{\bi}{\begin{itemize}}
\newcommand{\ei}{\end{itemize}}
\newcommand{\ben}{\begin{enumerate}}
\newcommand{\een}{\end{enumerate}}
\newcommand{\bd}{\begin{definition}}
\newcommand{\ed}{\end{definition}}
\def\beq{\begin{equation}}
\def\eeq{\end{equation}\noindent}
\def\beqn{\begin{eqnarray}}
\def\eeqn{\end{eqnarray} \noindent}
\def\beqnn{  \begin{eqnarray*}}
\def\eeqnn{\end{eqnarray*}  \noindent}
\def\bcase{  \begin{numcases}}
\def\ecase{\end{numcases}   \noindent}
\def\bsbcase{  \begin{subnumcases}}
\def\esbcase{\end{subnumcases}   \noindent}

\newenvironment{myproof}{\noindent{\em Proof:} \hspace*{1em}}{
    \hspace*{\fill} $\Box$ }
\newenvironment{proof_of}[1]{\noindent {\em Proof of #1: }}{\hspace*{\fill} $\Box$ }

\newcommand{\matplottc}[1]{               
        \unitlength .45truein
        \begin{center}
        \includegraphics{#1.ps}
        \end{picture}
        \end{center}
}

\def\psfancypar#1#2{\begingroup\def\par{\endgraf\endgroup\lineskiplimit=0pt}
               \setbox2=\hbox{\large\sc #2}
               \newdimen\tmpht \tmpht \ht2 \advance\tmpht by \baselineskip
               \font\hhuge=Times-Bold at \tmpht
               \setbox1=\hbox{{\hhuge #1}}
               \count7=\tmpht \count8=\ht1
               \divide\count8 by 1000 \divide\count7 by \count8
               \tmpht=.001\tmpht\multiply\tmpht by \count7
               \font\hhuge=Times-Bold at \tmpht
               \setbox1=\hbox{{\hhuge #1}}
               \noindent
                \hangindent1.05\wd1
               \hangafter=-2 {\hskip-\hangindent
               \lower1\ht1\hbox{\raise1.0\ht2\copy1}%
                \kern-0\wd1}\copy2\lineskiplimit=-1000pt}

\def\Kout{\setbox1=\hbox{\Huge\bf K}\hbox to
1.05\wd1{\hspace{.05\wd1}
}}
\def\Sout{\setbox1=\hbox{\Huge\bf S}\hbox to 1.05\wd1{\hspace{.05\wd1}
}}

\newcommand{\ranktest}{\mathsf{RankTest}}
\newcommand{\FindComponents}{\mathsf{FindMixtureComponents}}
\newcommand{\SpectDecomp}{\mathsf{SpecDecom}}
\newcommand{\TreeApprox}{\mathsf{ChowLiuTree}}
\newcommand{\MaxWtTree}{\mathsf{MaxWtTree}}
\newcommand{\eigvalue}{\mathsf{Eigenvalues}}
\DeclareMathOperator{\tree}{tree}
\DeclareMathOperator{\spect}{spect}

\DeclareMathOperator{\lspect}{local-spect}
\def\tilM{\widetilde{M}}

\def\cP{\breve{P}}

\def\R{\mathbb{R}}

\def\sphere{\mathbb{S}}

\renewcommand\v[1]{{\vec{#1}}}
\newcommand\dotp[1]{\langle #1 \rangle}
\def\t{{\scriptscriptstyle\top}}
\def\h{\widehat}

\def\e{{\vec{e}}}

\def\eps{\epsilon}
\def\veps{\varepsilon}

\DeclareMathOperator{\diag}{Diag}

\def\bs{\kern-2pt}

\begin{document}

\title{Learning High-Dimensional Mixtures of Graphical Models}

\author{%
Animashree Anandkumar \\ \url{a.anandkumar@uci.edu}\\
 UC Irvine
 \and%
  Daniel Hsu\\ \url{dahsu@microsoft.com}\\
 Microsoft Research New England\and
Furong Huang\\ \url{furongh@uci.edu}\\ UC Irvine \and
 Sham M.~Kakade\\ \url{skakade@microsoft.com}\\
  Microsoft Research New England
 }

\maketitle

\begin{abstract}We consider  unsupervised estimation of mixtures of discrete graphical models, where the class variable corresponding to the mixture components is hidden and each mixture component over the observed variables can have a potentially different Markov graph structure and parameters. We propose a novel approach for estimating the mixture components, and    our output is  a tree-mixture model which serves as a good approximation to the underlying graphical model mixture.
Our method is efficient when the union graph, which is the union of the Markov graphs of the mixture components, has sparse vertex separators between any  pair of observed variables. This includes tree mixtures and mixtures of bounded degree graphs. For such models, we prove that our method correctly recovers the union graph structure and the tree structures corresponding to maximum-likelihood tree approximations of the mixture components. The
  sample and computational complexities of our method scale as $\poly(p,   r)$,   for an $r$-component mixture of $p$-variate graphical models. We further extend our results to the case when the union graph has sparse local separators between any   pair of observed variables, such as mixtures of locally tree-like graphs,
 and the mixture components are in the regime of correlation decay.
\end{abstract}




\section{Introduction}

Graphical models offer a graph-based framework for representing multivariate distributions, where the structural and qualitative relationships between the variables are represented via a graph structure, while the parametric and quantitative relationships are represented via values assigned to different groups of nodes on the graph~\citep{Lauritzen:book}. Such a decoupling  is natural in a variety of contexts, including computer vision, financial modeling, and phylogenetics. Moreover, graphical models are amenable to efficient inference via distributed algorithms such as belief propagation~\citep{Wainwright&Jordan:08NOW}. Recent innovations have   made it feasible to train these models   with low computational and sample requirements  in high dimensions (see Section~\ref{sec:related} for a brief overview).

Simultaneously, much progress has been made  in analyzing  mixture models~\citep{lindsay1995mixture}. A mixture model can be thought of as selecting the  distribution of the manifest variables  from a fixed set, depending on the realization of a so-called choice variable, which is  latent or hidden. Mixture models have widespread applicability since they can account for changes in observed data based on hidden influences.
Recent works have provided provable guarantees for learning high-dimensional mixtures under a variety of settings  (See Section~\ref{sec:related}).

In this paper, we consider  mixtures of (undirected) graphical models, which combines the modeling power of the above two formulations. These models   can incorporate
{\em context-specific dependencies}, where the structural (and parametric) relationships among the observed variables can change depending on a hidden context. These models allow for parsimonious representation of high-dimensional data, while retaining the computational advantage of performing inference via belief propagation and its variants.
The current practice for learning mixtures of graphical models (and other mixture models) is based local-search heuristics such as expectation maximization (EM). However, EM  scales poorly in the number of  dimensions, suffers from  convergence issues, and   lacks  theoretical guarantees.

In this paper, we propose a novel approach for learning graphical model mixtures, which   offers a powerful alternative to EM. At the same time, we establish theoretical guarantees for our method for a wide class of models, which includes tree mixtures and mixtures over bounded degree graphs.  Previous theoretical guarantees  are mostly limited to  mixtures of product distributions  (see Section~\ref{sec:related}). These models are restrictive since they posit that the manifest variables are related to one another only via the latent choice variable, and have no direct dependence otherwise.   Our work is a significant generalization of these models,  and incorporates models such as tree mixtures and mixtures over bounded degree graphs.

Our approach aims to approximate the underlying graphical model mixture with   a tree-mixture model.   In our view, a tree-mixture approximation offers  good tradeoff between  data fitting and inferential complexity of the model. Tree mixtures are attractive since inference reduces to belief propagation on the component trees~\citep{meila2001learning}. Tree mixtures thus   present a middle ground between tree graphical models, which are too simplistic, and general graphical model mixtures, where   inference is not tractable, and our goal is to efficiently fit the observed data to a tree mixture model.

\subsection{Summary of Results}

We propose a novel method for learning discrete graphical mixture models. It
combines the techniques used in graphical model selection based on conditional independence tests,   and the spectral decomposition methods employed    for estimating the parameters of  mixtures of product distributions. Our method proceeds in three main stages: graph structure estimation, parameter estimation, and   tree approximation.




In the first stage, our algorithm estimates the union graph structure, corresponding to the union of the Markov graphs of the mixture components. We propose a rank criterion for classifying a node pair as neighbors or non-neighbors in the union graph of the mixture model, and can be viewed as a generalization of conditional-independence tests for graphical model selection~\citep{spirtes1995learning,AnandkumarTanWillsky:Ising11,Bresler&etal:Rand}. Our method is efficient when the union graph has sparse separators between any node pair, which holds for tree mixtures and mixtures of bounded degree graphs. The sample complexity of our algorithm is logarithmic in the number of nodes. Thus, our method learns the union graph structure of a graphical model mixture with similar guarantees as graphical model selection (i.e., when there is a single graphical model).

In the second stage, we use the union graph estimate $\hG_{\cup}$ to learn the pairwise marginals of the mixture components. Since the choice variable is hidden, this involves decomposition of the observed statistics into component models. We leverage on the spectral decomposition method developed for learning mixtures of product distributions~\citep{chang1996full,mossel2005learning,AnandkumarHsuKakade:COLT12}.
In a mixture of product distributions, the observed variables are conditionally independent given the hidden class variable.
We adapt this method to our setting as follows: we consider different triplets over the observed nodes  and  condition on suitable separator sets (in the union graph estimate $\hG_{\cup}$)   to obtain a series of mixtures of product distributions. Thus, we obtain estimates for pairwise marginals of each mixture component (and in principle,  higher order moments) under some natural non-degeneracy conditions.
In the final stage, we find the  best tree approximation to the estimated component marginals via the standard Chow-Liu algorithm~\citep{Chow&Liu:68IT}. The Chow-Liu algorithm produces a max-weight spanning tree   using the estimated pairwise mutual information as edge weights. We establish that our algorithm recovers the correct tree structure corresponding to maximum-likelihood tree approximation of each mixture component. In the special case, when the underlying distribution is a tree mixture, this implies that we can recover tree structures corresponding to all the mixture components.
The computational and sample complexities of our method scale as $\poly(p, r)$, where $p$ is the number of nodes and $r$ is the number of mixture components.


Recall that the success of our method relies on the presence of sparse vertex separators between node pairs in the union graph, i.e., the union of Markov graphs of the mixture components. This includes tree mixtures and mixtures of bounded degree graphs. We extend our methods and analysis the a larger family of models, where the union graph
has sparse local separators~\citep{AnandkumarTanWillsky:Ising11}, which is a weaker criterion. This family  includes  locally tree-like graphs (including sparse random graphs),  and augmented graphs (e.g. small-world graphs where there is a local and a global graph). The criterion of sparse local separation significantly widens the scope, and we prove that our methods succeed in these models, when the mixture components are in the regime of correlation decay~\citep{AnandkumarTanWillsky:Ising11}. The sample and computational complexities are significantly improved for this class, since it only depends on  the size of local separators (while previously it depended on the size of exact separators).


Our proof techniques involve establishing the correctness of our algorithm (under exact statistics). The sample analysis involves careful use of spectral perturbation bounds to guarantee success in finding the mixture components. In addition, for the setting with sparse local separators, we incorporate the correlation decay rate functions of the component models to quantify the additional distortion introduced due to the use of local separators as opposed to exact separators. One caveat of our method is that we require the dimension of the node variables $d$ to be larger than the number of mixture components $r$. In principle, this limitation can be overcome if we consider larger (fixed) groups of nodes and implement our method. Another limitation is that we require full rank views of the latent factor for our method to succeed. However, this is also a requirement imposed for learning mixtures of product distributions. Moreover,  it is  known that learning singular models, i.e., those which do not satisfy  the above rank condition, is at least as hard as learning parity with noise, which is conjectured to be computationally hard~\citep{mossel2005learning}. Another restriction is that we require the presence of an observed variable, which is conditionally independent of all the other variables, given the latent choice variable.  However, note that this is significantly weaker than the case of product mixture models, where all the observed variables are required to be conditionally independent given the latent factor.
To the best of our knowledge, our work is the first to provide provable guarantees for learning non-trivial graphical mixture models (which are not mixtures of product distributions), and we believe that it significantly advances the scope, both on theoretical and practical fronts.

\subsection{Related Work}\label{sec:related}

Our work lies at the intersection of learning mixture models and graphical model selection. We outline related works in both these areas.

\paragraph{Overview of Mixture Models: }
Mixture models have been extensively studied~\citep{lindsay1995mixture} and are employed in a variety of applications. More recently, the focus has been on learning mixture models in high dimensions.  There are a number of recent works dealing with estimation of high-dimensional Gaussian mixtures, starting from the work of~\citet{dasgupta1999learning} for learning well-separated components, and most recently  by~\citep{belkin2010polynomial,moitra2010settling}, in a long line of works. These works provide guarantees on recovery under various separation constraints between the mixture components and/or have computational and sample complexities growing exponentially in the number of mixture components $r$.
In contrast, the so-called spectral methods have both computational and sample complexities scaling only polynomially in the number of components, and do not impose stringent separation constraints, and we outline below.

\paragraph{Spectral Methods for Mixtures of Product Distributions: }
The classical mixture model over product distributions  consists of multivariate distributions with a single  latent variable $H$ and the observed variables are conditionally independent under each state of the latent  variable~\citep{lazarsfeld68}.    Hierarchical latent class (HLC) models~\citep{zhang04,ZhangJMLR04,Che08} generalize these models by allowing for multiple latent variables.
Spectral methods were first employed  for learning discrete (hierarchical) mixtures of product distributions~\citep{chang1996full,mossel2005learning,hsu2008spectral}  and have   been recently extended  for learning general multiview mixtures~\citep{AnandkumarHsuKakade:COLT12}. The method   is based on triplet and pairwise statistics of observed variables and we build on these methods in our work. Note that our setting is {\em not} a mixture of product distributions, and thus, these methods are not directly applicable.


\paragraph{Graphical Model Selection: }Graphical model selection is a well studied problem starting from the seminal work of~\citet{Chow&Liu:68IT} for finding the best tree approximation of a graphical model. They established that maximum likelihood estimation reduces to a maximum weight spanning tree problem where the edge weights are given by empirical mutual information. However, the problem becomes more challenging when either some of the nodes are hidden (i.e., latent tree models) or we are interested in estimating loopy graphs.     Learning the structure of  latent tree models has   been studied extensively, mainly in the context of phylogenetics~\citep{Durbin}.   Efficient algorithms with provable performance guarantees are  available, e.g. \citep{erdos99,daskalakis06,Choi&etal:10JMLR,AnandkumarEtal:spectral}. Works on high-dimensional loopy graphical model selection are more recent. The approaches can be classified into mainly  two groups:  non-convex local  approaches~\citep{Anandkumar:girth12,AnandkumarTanWillsky:Ising11,Jalali:greedy,Bresler&etal:Rand,Sanghavi&etal:Allerton10} and those based on convex optimization~\citep{Mei06,Ravikumar&etal:08Arxiv,Ravikumar&etal:08Stat,Chandrasekaran:10latent}. There is also some recent work on learning conditional models, e.g.~\citep{guo2011joint}.    However, these works are not directly applicable for learning mixtures of graphical models.

\paragraph{Mixtures of Graphical Models: }Works on learning mixtures of graphical models (other than mixtures of product distributions) are fewer, and mostly focus on tree mixtures.     The works of~\citet{meila2001learning} and \citet{kumar2009learning} consider EM-based approaches for learning tree mixtures, ~\citet{thiesson1999computationally} extend the  approach to learn   mixtures of graphical models on directed acyclic graphs (DAG), termed as Bayesian multinets by~\citet{geiger1996knowledge}, using the  Cheeseman-Stutz asymptotic approximation and~\citet{Armstrong:2009} consider a  Bayesian approach by assigning a prior to decomposable graphs. However, these approaches do not  have any theoretical guarantees.

Recently,~\citet{mossel2011phylogenetic} consider structure learning of latent tree mixtures and provide conditions under which they can be successfully recovered.  Note that this model   can be thought of as a hierarchical mixture of product distributions, where the hierarchy changes according to the realization of the choice variable.   Our setting differs substantially from this work.  \citet{mossel2011phylogenetic} require that  the component latent trees of the mixture  be very different, in order for the quartet tests to distinguish them (roughly), and establish that  a uniform selection of trees  will ensure this condition. On the other hand, we impose no such restriction and allow for graphs of different components to be same/different (although our algorithm is efficient when the overlap between the component graphs is more).
Moreover, we allow for loopy graphs while  \citet{mossel2011phylogenetic} restrict to learning latent tree mixtures. However,~\citet{mossel2011phylogenetic} do allow for latent variables on the tree, while we assume that all variables to be observed (except for the latent choice variable).~\citet{mossel2011phylogenetic} consider only structure learning, while we consider both structure and parameter estimations.~\citet{mossel2011phylogenetic} limit to finite number of mixtures $r=O(1)$, while we allow for $r$ to scale with the number of variables $p$. As such, the two methods operate in significantly different settings.

\section{System Model}

\subsection{Graphical Models}

We first introduce the concept of a graphical model and then discuss mixture models. A   graphical model is a family of  multivariate distributions   Markov on a given undirected graph~\citep{Lauritzen:book}.
In a discrete graphical model, each node in the graph $v\in V$ is associated with a random variable $Y_v$ taking value in a finite set $\Yc$   and let $d:=|\Yc|$ denote the cardinality of the set. The set of edges\footnote{We use notations $E$ and $G$ interchangeably to denote the set of edges.}   $E\subset\binom{V}{2}$ captures the set of conditional-independence relationships among the random variables.   We say that a vector of random variables $\bfY:=(Y_1,\ldots, Y_p)$ with a joint  probability mass function (pmf) $P$ is Markov on the graph $G$ if the {\em local Markov property}
\begin{equation}
P(y_v|\bfy_{\nbd(i)}) = P(y_v|\bfy_{V\setminus v})
\end{equation}
holds for all nodes $v \in V$, where $\nbd(v)$ denotes the open neighborhood of $v$ (i.e., not including $v$).
More generally, we say that  $P$ satisfies the {\em global Markov property} for all disjoint  sets $A,B\subset V$
\beq P(\bfy_A, \bfy_B|\bfy_{\Sc(A, B;G)}) = P(\bfy_A|\bfy_{\Sc(A, B;G)}) P(\bfy_B|\bfy_{\Sc(A, B;G)}),\quad \forall A,B\subset V:\nbd[A]\cap \nbd[B]=\emptyset.\eeq where the   set ${\Sc(A, B;G)}$ is a {\em node separator}\footnote{A set ${\Sc(A, B;G)}\subset V$ is a separator of sets $A$ and $B$ if the removal of nodes in ${\Sc(A, B;G)}$ separates $A$ and $B$ into distinct components.}between $A$ and $B$, and $\nbd[A]$ denotes the closed  neighborhood of $A$ (i.e., including $A$). The global and local Markov properties are equivalent under the {\em positivity condition}, given by $P(\bfy)>0$, for all $\bfy\in \Yc^p$~\citep{Lauritzen:book}. Henceforth, we say that a graphical model satisfies Markov property with respect to a graph, if it satisfies the global Markov property.

The Hammersley-Clifford theorem~\citep{Bremaud:book} states that under the positivity condition, a distribution $P$  satisfies the Markov property according to a graph $G$ iff. it factorizes according to the cliques of $G$,
\beq P(\bfy) = \frac{1}{Z}\exp\left(\sum_{c\in \Cc}\Psi_c(\bfy_c)\right),\eeq where $\Cc$ is the set of cliques of $G$ and $\bfy_c$ is the set of random variables on clique $c$.   The quantity $Z$ is known as the {\em partition function} and serves to normalize the probability distribution. The functions $\Psi_c$ are known as {\em potential} functions. We will assume positivity of the graphical models under consideration, but otherwise allow for general potentials (including higher order potentials).

\subsection{Mixtures of  Graphical Models}

In this paper, we consider mixtures of discrete graphical models. Let $H$ denote the discrete  hidden choice variable corresponding to the selection of a different components of the mixture, taking values in $[r]:=\{1, \ldots, r\}$ and let $\bfY$ denote the observed variables of the mixture.    Denote $\pibf_H:=[P(H=h)]^\top_h$ as the probability vector of the mixing weights and   $G_h$ as the Markov graph of the distribution $P(\bfy|H=h)$.

Our goal is to learn the mixture of graphical models, given $n$ i.i.d.\ samples $\bfy^n=[\bfy_1, \ldots, \bfy_n]^\top$ drawn from a $p$-variate joint distribution $P(\bfy)$ of the mixture model, where each variable is a $d$-dimensional discrete variable. The component Markov graphs $\{G_h\}_h$ corresponding to models $\{P(\bfy|H=h)\}_h$ are assumed to be unknown.  Moreover,  the variable $H$ is latent and thus, we do not a priori know the mixture component from which a sample is drawn.  This implies that  we cannot directly apply the previous methods  designed for graphical model selection. A major challenge is thus being able to decompose the observed statistics into the mixture components.

We now propose a method for learning the mixture components given $n$ i.i.d.\ samples $\bfy^n$ drawn from  a graphical mixture model $P(\bfy)$. Our method proceeds in three main stages.  First, we estimate the graph $G_{\cup}:=\cup_{h=1}^r G_h$, which is the union of the Markov graphs of the mixture. This is accomplished via a series of rank tests. Note that in the special case when $G_h \equiv G_{\cup}$, this   gives the graph estimates of the component models. We then use the graph estimate $\hG_{\cup}$ to obtain the pairwise marginals of the respective mixture components via a spectral decomposition method. Finally, we use the Chow-Liu algorithm to obtain tree approximations $\{T_h\}_h$ of the individual mixture components\footnote{Our method can also be adapted to estimating the component Markov graphs $\{G_h\}_h$ and we outline it and other extensions  in Appendix~\ref{sec:graphest}.}.

\section{Estimation of the Union of Component Graphs}

\paragraph{Notation:} Our learning method will be based on the estimates of  probability matrices. For any two nodes $u,v\in V$ and any set $S\subset V\setminus \{u,v\}$, denote the joint probability matrix \beq \label{eqn:M}M_{u,v,\{S;k\}}:=[P(Y_u=i, Y_v=j,\bfY_S=k)]_{i,j},\quad k \in \Yc^{|S|}.\eeq
Let $\hM^n_{u,v,\{S;k\}}$ denote the corresponding estimated matrices using samples $\bfy^n$\beq \hM^n_{u,v,\{S;k\}}:= [\hP^n(Y_u=i, Y_v=j,\bfY_S=k)]_{i,j},\label{eqn:Mest}\eeq  where $\hP^n$ denotes the empirical   probability distribution, computed using $n$ samples. We consider sets $S$ satisfying $|S|\leq \eta$, where $\eta$   depends on the graph family under consideration. Thus, our method is based on $(\eta+2)^{\tha}$ order statistics of the observed variables.

\paragraph{Intuitions: }We  provide some intuitions and properties of  the union   graph
$G_{\cup}=\cup_{h=1}^r G_h$, where $G_h$ is the Markov graph corresponding to component $H=h$. Note that $G_{\cup}$ is different from the Markov graph corresponding to the marginalized model $P(\bfy)$ (with latent choice variable $H$ marginalized out). Yet, $G_{\cup}$ represents some natural Markov properties with respect to the observed statistics. We first establish the simple result that the union graph $G_{\cup}$  satisfies Markov property in each mixture component. Recall that $\Sc(u,v;G_{\cup})$ denotes a vertex separator between nodes $u$ and $v$ in $G_{\cup}$, i.e., its removal disconnects $u$ and $v$ in $G_{\cup}$.

\begin{fact}[Markov Property of $G_{\cup}$]\label{fact:condindepunion}For any two nodes $u,v \in V$ such that $(u,v)\notin G_{\cup}$,   \beq\label{eqn:condindepunion} Y_u\indep Y_v|\bfY_{S}, H, \quad S:=\Sc(u,v;G_{\cup}).\eeq\end{fact}

\bprf  The set  $S:=\Sc(u,v;G_{\cup})$ is also a vertex separator for $u$ and $v$ in each of the component graphs $G_h$. This is because removal of $S$ disconnects $u$ and $v$ in each $G_h$. Thus, we have Markov property in each component: $Y_u\indep Y_v|\bfY_{S}, \{H=h\}$, for $h \in [r]$, and the above result follows.\eprf

Thus, the above observation implies that the conditional independence relationships of each mixture component are satisfied  on the union graph $G_{\cup}$ conditioned on the latent factor $H$. The above result can be exploited to obtain union graph estimate as follows: two nodes $u,v$ are not neighbors in $G_{\cup}$ if a separator set $S$ can be found which results in conditional independence, as in \eqref{eqn:condindepunion}. The main challenge is indeed that the variable $H$ is not observed and thus, conditional independence cannot be directly inferred via observed statistics. However, the effect of $H$ on the observed statistics can be quantified as follows:

\begin{lemma}[Rank Property]\label{lemma:sep}Given an $r$-component mixture of graphical models with $G_{\cup}=\cup_{h=1}^r G_h$, for any $u,v\in V$ such that $(u,v)\notin G_{\cup}$ and $S:=\Sc(u,v;G_{\cup})$, the probability matrix $M_{u,v,\{S;k\}}:=[P[Y_u=i, Y_v=j,\bfY_S=k]]_{i,j}$ has rank at most $r$ for any $k \in \Yc^{|S|}$.\end{lemma}

 \bprf From Fact~\ref{fact:condindepunion},    $G_{\cup}$ satisfies Markov property conditioned on the latent factor $H$, \beq Y_u\indep Y_v|\bfY_S, H, \quad \forall\,(u,v)\notin G_{\cup}.\label{eqn:cond}\eeq
 This implies that  \beq \label{eqn:uv}M_{u,v,\{S;k\}} =   M_{u|H,\{S;k\}} \Diag(\pibf_{H|\{S;k\}}) M^\top_{v|H,\{S;k\}} P(\bfY_S=k),\eeq where $M_{u|H, \{S;k\}}:=[P[Y_u=i| H=j,\bfY_S=k]]_{i,j}$  and similarly $M_{v|H,\{S;k\}}$ is defined. $\Diag(\pibf_{H|S;k})$ is the diagonal matrix with entries $\pibf_{H|\{S;k\}}:= [P(H=i|\bfY_S=k)]_i$. Thus, we have $ \rank(M_{u,v,\{S;k\}})$ is at most $r.$ \eprf

Thus, the effect of marginalizing the choice variable $H$ is seen in the rank of the observed probability matrices $M_{u,v,\{S;k\}}$. Thus, when $u$ and $v$ are non-neighbors in $G_{\cup}$, a separator set $S$ can be found such that the rank of $M_{u,v,\{S;k\}}$ is at most $r$. In order to use this result as a criterion for inferring neighbors in $G_{\cup}$, we require that the rank of $M_{u,v,\{S;k\}}$ for any neighbors $(u,v)\in G_{\cup}$ be strictly larger than $r$. This requires the dimension of each node variable $d>r$. We discuss in detail the set of sufficient conditions for correctly recovering $G_{\cup}$ in Section~\ref{sec:assumptions-rank-exact}.

\paragraph{Tractable Graph Families: }Another obstacle in using Lemma~\ref{lemma:sep} to estimate graph $G_{\cup}$ is computational: the search for separators $S$ for any node pair $u,v\in V$ is exponential in $|V|:=p$ if no further constraints are imposed. We consider graph families where a vertex separator can be found  for any $(u,v)\notin G_{\cup}$ with size at most $\eta$\[ |\Sc(u,v;G_{\cup})|\leq \eta, \quad \forall (u,v)\notin G_{\cup}.\] There are many natural families where $\eta$ is small:\ben \item
If $G_{\cup}$ is trivial (i.e., no edges) then $\eta=0$, we have a mixture of product distributions. \item When $G_{\cup}$ is a tree, i.e.,  we have a mixture model Markov on the same tree, then $\eta=1$, since there is a unique path between any two nodes on a tree. \item For an arbitrary $r$-component tree mixture, $G_{\cup}= \cup_h T_h$ where each component is a tree distribution, we have that   $\eta \leq r$ (since  for any node pair, there is a unique path in each of the $r$ trees $\{T_h\}$, and separating the node pair in each $T_h$ also separates them on $G_{\cup}$).
\item For an arbitrary mixture of bounded degree graphs, we have $\eta\leq \sum_{h\in [r]}\Delta_h$, where $\Delta_h$ is the maximum degree  in $G_h$, i.e., the Markov graph corresponding to component $\{H=h\}$.\een In general,  $\eta$ depends on the respective bounds $\eta_h$ for the component graphs $G_h$, as well as the extent of their overlap. In the worst case, $\eta$ can be as high as $\sum_{h\in [r]}\eta_h$, while in the special case when $G_h \equiv G_{\cup}$, the bound  remains the same $\eta_h \equiv \eta$.
Note that for a general graph $G_{\cup}$ with  {\em treewidth} $\tw(G_{\cup})$  and maximum degree $\Delta(G_{\cup})$, we have that $\eta\leq \min(\Delta(G_{\cup}),\tw(G_{\cup}))$. Thus, a wide family of models give rise to union graph with small $\eta$, including tree mixtures and mixtures over bounded degree graphs.

We establish in the sequel that the computational and sample complexities of our learning method  scale exponentially in $\eta$. Thus, our algorithm is suitable for graphs $G_{\cup}$ with small $\eta$. In Section~\ref{sec:corrdecay},  we relax the requirement of exact separation to that of local separation. A larger class of graphs satisfy the local separation property including mixtures of locally tree-like graphs.

\paragraph{Rank Test: }
We propose $\ranktest(\bfy^n;\xi_{n,p},\eta,r)$ in Algorithm~\ref{algo:ranktest}  for structure estimation of $G_{\cup}:=\cup_{h=1}^r G_h$, the  union  Markov graph of an $r$-component mixture. The method is based on a search for potential separators $S$ between any two given nodes $u, v\in V$, based on the effective rank\footnote{The effective rank  is given by the number of singular values above a given   threshold $\xi$.  }of $\hM^n_{u,v,\{S;k\}}$: if the effective rank is $r$ or less, then $u$ and $v$ are declared as non-neighbors (and set $S$ as their separator). If no such sets are found, they are declared as neighbors. Thus, the method involves searching for separators for each node pair $u,v\in V$, by considering all sets $S\subset V\setminus \{u,v\}$ satisfying $|S|\leq \eta$.
The computational complexity of this procedure is $O(p^{\eta+2}d^3)$, where $d$ is the dimension of each node variable $Y_i$, for $i \in V$ and $p$ is the number of nodes. This is because  the number of rank tests performed is $O(p^{\eta+2})$ over all node pairs and conditioning sets; each rank tests has $O(d^3)$ complexity since it involves performing singular value decomposition (SVD) of a $d\times d$ matrix.

\begin{algorithm}[t]\begin{algorithmic}
\caption{$\hG^n_{\cup}=\ranktest(\bfy^n;\xi_{n,p},\eta,r)$
for estimating $G_{\cup}:=\cup_{h= 1}^r G_h$ of an $r$-component mixture using   $\bfy^n$ samples, where $\eta$ is the bound on size of vertex separators between any node pair in $G_{\cup}$ and $\xi_{n,p}$ is a threshold on the singular values.}\label{algo:ranktest}
\STATE  $\rank(A;\xi)$ denotes the effective rank of matrix $A$, i.e., number of singular values more than $\xi$. $\hM^n_{u,v,\{S;k\}}:=[\hP^n(Y_u=i, Y_v=j,\bfY_S=k)]_{i,j}$ is the empirical estimate   computed using $n$ i.i.d.\ samples $\bfy^n$. Initialize $\hG_{\cup}^n= (V, \emptyset)$.\STATE
For each $u,v \in V$, estimate $\hM^n_{u,v,\{S;k\}}$ from $\bfy^n$, if
\beq\label{eqn:sep}\min_{\substack{S\subset V\setminus\{u,v\}\\ |S|\leq \eta}}\max_{k \in \Yc^{|S|}}\rank(\hM^n_{u,v,\{S;k\}};\xi_{n,p})> r,\eeq then  add $(u,v)$ to $\hG_{\cup}^n$.
\end{algorithmic}
\end{algorithm}

\subsection{Results for the Rank Test}
\subsubsection{Conditions for the Success of Rank Tests}\label{sec:assumptions-rank-exact}
The following assumptions are made for the $\ranktest$ proposed in Algorithm~\ref{algo:ranktest} to succeed  under the PAC formulation.

\ben

\item[(A1)] {\bf Number of Mixture Components: } The number of components $r$ of the mixture model and dimension $d$ of each node variable  satisfy \beq d> r.\eeq The mixing weights of the latent factor $H$ are assumed to be strictly positive\[\pi_H(h):= P(H=h)>0, \quad \forall\,\, h\in[ r].\]

\item[(A2)] {\bf Constraints on Graph Structure: }Recall that  $G_{\cup}=\cup_{h=1}^r G_h$ denotes the union of the  graphs of the components and that $\eta$ denotes the bound on the size of the minimal separator set for   any two (non-neighboring) nodes in  $G_{\cup}$. We assume that\[ |\Sc(u,v;G_{\cup})|\leq \eta=O(1), \quad \forall (u,v)\notin G_{\cup}.\]
In Section~\ref{sec:corrdecay}, we relax the strict separation constraint to a local separation constraint in the regime of correlation decay, where $\eta$ refers to the bound on the size of local separators between any two non-neighbor nodes in the graph.

\item[(A3)] {\bf Rank Condition: }We assume that the matrix $M_{u,v,\{S;k\}}$ in \eqref{eqn:M} has rank strictly greater than $r$ when the nodes $u$ and $v$  are neighbors in graph $G_{\cup}=\cup_{h=1}^r G_h$ and the set satisfies $|S|\leq \eta$. Let $\rho_{\min}$ denote  \beq \label{eqn:rhomin}\rho_{\min}:= \min_{\substack{(u,v)\in G_{\cup}, |S|\leq \eta \\ S\subset V\setminus \{u,v\}}} \max_{k\in \Yc^{|S|}}\sigma_{r+1}\left(M_{u,v,\{S;k\}}\right)>0,\eeq where $\sigma_{r+1}(\cdot)$ denotes the $(r+1)^{\tha}$  singular value, when the singular values are arranged in the descending order $\sigma_1(\cdot)\geq \sigma_2(\cdot)\geq\ldots \sigma_d(\cdot)$.

\item[(A4)] {\bf Choice of threshold $\xi$: }For $\ranktest$ in Algorithm~\ref{algo:ranktest}, the threshold $\xi$ is chosen as \[ \xi:= \frac{\rho_{\min}}{2}. \]

\item[(A5)] {\bf Number of Samples: }Given  $\delta\in (0,1)$, the number of samples $n$ satisfies \beq n > n_{\rank}(\delta;p):= \max\left(\frac{1}{t^2}\left(2 \log p + \log \delta^{-1} +\log 2\right), \left(\frac{2}{\rho_{\min}-t}\right)^2\right),\eeq for some $t\in (0,\rho_{\min})$ (e.g. $t=\rho_{\min}/2$,) where $p$ is the number of nodes and $\rho_{\min}$ is given by \eqref{eqn:rhomin}.

\een

Assumption (A1) relates the number of components to the dimension of the sample space of the variables. Note that we allow for the number of components $r$ to grow with the number of nodes $p$, as long as the cardinality of the sample space of each variable $d$ is also large enough.  In principle, this assumption can be removed by considering grouping the nodes together and performing rank tests on the groups.
Assumption (A2) imposes constraints on the graph structure $G_{\cup}$, formed by the union of the component graphs. The  bound $\eta$ on the separator sets for node pairs in $G_{\cup}$ is a crucial parameter and the complexity of learning (both sample and computational) depends on it. We relax the assumption of separator bound to a criterion of local separation in Section~\ref{sec:corrdecay}.
Assumption (A3) is required for the success of   rank tests to distinguish neighbors and non-neighbors in graph $G_{\cup}$. It rules out the presence of spurious  low rank  matrices between neighboring nodes in $G_{\cup}$ (for instance, when the nodes are marginally independent or when the distribution is degenerate). Assumption (A4) provides a natural  threshold on the singular values in the rank test. In Section~\ref{sec:corrdecay}, we modify the threshold to also account for distortion due to approximate vertex separation, in contrast to the setting of exact separation considered in this section.
 (A5) provides the finite sample complexity bound.

%



\subsubsection{Result on Rank Tests}

We now provide the result on the success of recovering the graph $G_{\cup}:=\cup_{h=1}^r G_h$.

\bt[Success of Rank Tests]\label{thm:ranktest} The $\ranktest(\bfy^n;\xi,\eta, r)$ outputs the correct graph $G_{\cup}:=\cup_{h=1}^r G_h$, which is the union of the component Markov graphs, under the assumptions (A1)--(A5) with probability at least $1-\delta$.\et

\bprf The proof is given in Appendix~\ref{proof:ranktest}. \eprf

A special case of the above result is graphical model selection, where there is a single graphical model $(r=1)$ and we are interested in estimating its graph structure.

\begin{corollary}[Application to Graphical Model Selection] The $\ranktest(\bfy^n;\xi,\eta, 1)$ outputs the correct Markov graph $G$, given $n$ i.i.d. samples $\bfy^n$, under the assumptions\footnote{When $r=1$, there is no latent factor, and the assumption $d>r$ in (A1) is trivially satisfied for all discrete random variables.} (A2)--(A5) with probability at least $1-\delta$.\end{corollary}

\noindent{\bf Remarks: }Thus, the rank test is also applicable  for graphical model selection. Previous works (see Section~\ref{sec:related}) have proposed tests based on conditional independence, using either conditional mutual information or conditional variation distances, see~\cite{Bresler&etal:Rand,AnandkumarTanWillsky:Ising11}. The rank test above is thus an alternative test for conditional independence.  In addition, it extends naturally to estimation of union graph structure of mixture components.

\section{Parameter Estimation of Mixture Components}

The rank test proposed in the previous section  is a tractable procedure for estimating the graph $G_{\cup}:=\cup_{h=1}^r G_h$, which is the union of the component graphs of a mixture of graphical models. However, except in the special case when $G_h \equiv G_{\cup}$,  the knowledge of $\hG_{\cup}^n$ is not very useful by itself, since  we do not know the nature of the different components of the mixture. In this section, we propose the use of spectral decomposition tests to find the various mixture components.

\subsubsection{Spectral Decomposition  for Mixture of Product Distributions}
The spectral decomposition methods, first proposed by~\citet{chang1996full}, and later generalized by~\citet{mossel2005learning} and~\citet{hsu2008spectral}, and recently by~\citet{AnandkumarHsuKakade:COLT12}, are applicable for mixtures of product distributions. We illustrate the method below via a simple example.

\begin{wrapfigure}{r}{0.8in}
\bc\bp\psfrag{u}[l]{$u$}\psfrag{H}[r]{\tcdkg{$H$}}
\psfrag{v}[l]{$v$}\psfrag{w}[l]{$w$}\psfrag{S}[l]{\tcv{$S$}}
\fbox{\includegraphics[width=0.8in,height=0.8in]{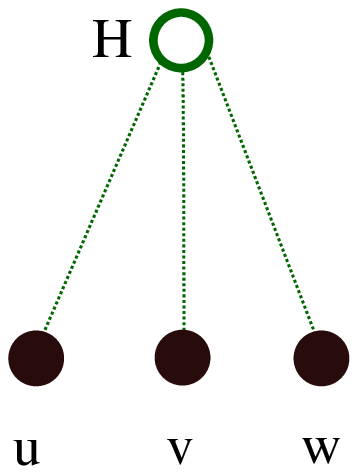}}\ep\ec
\end{wrapfigure}Consider the simple case of three observed variables $\{Y_u,Y_v,Y_w\}$, where   a latent factor $H$  separates    them, i.e., the observed variables are conditionally independent given $H$\[ Y_u\indep Y_v\indep Y_w|H.\] This implies that the Markov graphs $\{G_h\}_{h\in [ r]}$ of the component models $\{P(Y_u, Y_v,Y_w|H=h)\}_{h\in [ r]}$ are trivial (i.e., have no edges) and thus forms a special case of our setting.

We now give an overview of the spectral decomposition method.
It proceeds by considering pairwise and triplet statistics of $Y_u, Y_v, Y_w$.  Denote  $M_{u|H}:=[P(Y_u=i|H=j)]_{i,j}$, and similarly for $M_{v|H}, M_{w|H}$ and assume that they are full rank. Denote the   probability matrices $M_{u,v}:= [P(Y_u =  i, Y_v=j)]_{i,j}$ and $M_{u,v,\{w;k\}}:= [P(Y_u =  i, Y_v=j, Y_w=k)]_{i,j}$. The parameters (i.e., matrices $M_{u|H}, M_{v|H}, M_{w|H}$) can be estimated as:

\begin{lemma}[Mixture of Product Distributions]\label{lemma:lcm}For the latent variable model  $Y_u \indep Y_v \indep Y_w|H$, when the conditional probability matrices $M_{u|H}, M_{v|H}, M_{w|H}$ have rank $d$,
let  $\lambdabf^{(k)}=[\lambda^{(k)}_1, \ldots, \lambda^{(k)}_d]^\top$ be the column vector with the $d$ eigenvalues given by
\beq\label{eqn:eig} \lambdabf^{(k)}:=\eigvalue\left( M_{u,v,\{w;k\}} M_{u,v}^{-1}\right), \quad k \in \Yc.\eeq
Let $\Lambda:= [\lambdabf^{(1)}| \lambdabf^{(2)}|\ldots| \lambdabf^{(d)}]$ be the  matrix  where the $k^{\tha}$ column corresponds to $\lambdabf^{(k)}$ from above. We have that \beq\label{eqn:spec} M_{w|H}:=[P(Y_w=i|H=j)]_{i,j} =\Lambda^\top.\eeq \end{lemma}

\bprf A more general result is proven in Appendix~\ref{proof:spectexact}.\eprf

Thus, we have a procedure for recovering the conditional probabilities of the observed variables conditioned on the latent factor. Using these parameters, we can also recover the mixing weights $\pibf_H:= [P(H=i)]^\top_i$ using the relationship \[ M_{u,v}= M_{u|H} \Diag(\pibf_H) M_{v|H}^\top,\] where $\Diag(\pibf_H)$ is the diagonal matrix with $\pibf_H$ as the diagonal elements.

Thus, if we have a general product distribution mixture over nodes in $V$, we can learn the parameters by performing the above spectral decomposition over different triplets $\{u,v,w\}$.
However, an obstacle remains: spectral decomposition over different  triplets $\{u,v,w\}$   results in different permutations of the labels of the hidden variable $H$. To overcome this,  note that any two triplets $(u,v,w)$ and $(u,v',w')$ share the same set of eigenvectors in \eqref{eqn:eig} when the   ``left'' node $u$  is the same.
Thus, if we consider   a fixed node $u_*\in V$ as the ``left'' node  and use a fixed matrix to  diagonalize \eqref{eqn:eig} for all triplets, we obtain a consistent ordering of the hidden  labels over all triplet decompositions. Thus, we can learn a general product distribution mixture using only third-order statistics.

\subsubsection{Spectral Decomposition for Learning Graphical Model Mixtures}

\begin{figure}[h]
\centering{
\bp\psfrag{w1}[l]{\tccyan{$w$}}\psfrag{w2}[l]{\tccyan{$w'$}}
\psfrag{c}[c]{$\tcr{v}$}
\psfrag{H}[l]{\tcdkg{$H$}}\psfrag{G}[c]{$G_{\cup}$}\psfrag{u}[l]{\tcb{$u$}}
\psfrag{S}[c]{$\tcv{S}$}
\includegraphics[width=2in]{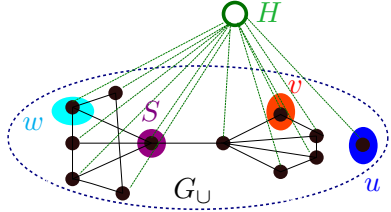}
\ep}
\caption{By conditioning on the separator set $S$ on the union graph $G_{\cup}$, we have a mixture of product distribution with respect to nodes $\{u,v,w\}$, i.e., $Y_u\indep Y_v\indep Y_w|\bfY_S, H$.}\label{fig:mixtureseparation}\end{figure}

We now adapt the above method for learning more general graphical model mixtures. We first make a simple observation on how to obtain mixtures of product distributions by considering separators on the union graph $G_{\cup}$. For any three nodes $u,v,w\in V$, which are not neighbors on $G_{\cup}$, let $S_{uvw}$ denote a {\em multiway} vertex separator, i.e., the removal of nodes in $S_{uvw}$ disconnects $u,v$ and $w$ in $G_{\cup}$. On lines of Fact~\ref{fact:condindepunion},  \beq Y_u \indep Y_v \indep Y_w | \bfY_{S_{uvw}}, H, \quad \forall u,v,w: (u,v), (v,w), (w,u)\notin G_{\cup}.\eeq Thus, by fixing the configuration of nodes in $S_{uvw}$, we obtain a product distribution mixture over $\{u,v,w\}$. If the previously proposed rank test is successful in estimating $G_{\cup}$, then we possess correct knowledge of the separators $S_{uvw}$. In this case, we can obtain estimates $\{P(Y_w|\bfY_{S_{uvw}}=k, H=h)\}_h$ by fixing the nodes in $S_{uvw}$ to $k$ and using the  spectral decomposition described in Lemma~\ref{lemma:lcm}, and the procedure can be repeated over different triplets $\{u,v,w\}$. See
Fig.\ref{fig:mixtureseparation}.

An obstacle remains, \viz the permutation of hidden labels over different triplet decompositions $\{u,v,w\}$. In case of product distribution mixture, as discussed previously, this is resolved by fixing the ``left'' node in the triplet to some $u_* \in V$ and using the same matrix for diagonalization over different triplets. However, an additional complication arises when we consider graphical model mixtures, where conditioning over separators is required. We require that the permutation of the hidden labels be unchanged upon conditioning over different values of variables in the separator set $S_{u_*vw}$. This holds when  the separator set $S_{u_*vw}$ has no effect on node $u_*$, i.e.,  we require that \beq \label{eqn:isolated}\exists u_* \in V, \st \quad Y_{u_*}\indep \bfY_{V\setminus u_*} | H,\eeq which implies that $u_*$ is isolated from all other nodes  in graph $G_{\cup}$.

Condition  \eqref{eqn:isolated} is required to hold for identifiability if we only operate on statistics over different triplets (along with their separator sets). In other words, if we resort to operations over only low order statistics, we require additional conditions such as \eqref{eqn:isolated} for identifiability. However, our setting is a significant generalization over the mixtures of product distributions, where \eqref{eqn:isolated} is required to hold for all nodes.

Finally, since our goal is  to estimate pairwise marginals of the mixture components,  in place of node $w$ in the triplet $\{u,v,w\}$ in Lemma~\ref{lemma:lcm}, we need to consider a node pair $a,b\in V$. The general algorithm allows the variables in the triplet to have different dimensions, see~\cite{AnandkumarHsuKakade:COLT12} for details. Thus, we   obtain estimates of the pairwise marginals of the mixture components. The computational complexity of the procedure scales  as $O(p^2 d^{\eta+6} r)$, where $p$ is the number of nodes, $d$ is the cardinality of each node variable and $\eta$ is the bound on separator sets.  For details on implementation of the spectral method, see Appendix~\ref{sec:specdetails}.

\subsection{Results for Spectral Decomposition}\label{sec:exactsepspectral}

\subsubsection{Assumptions}
In addition to the assumptions (A1)--(A5) in Section~\ref{sec:assumptions-rank-exact}, we impose the following constraints to guarantee the success of estimating the various   mixture components.

\ben

\item[(A6)]{\bf Full Rank Views of the Latent Factor: }For each node pair $a,b\in V$, and any subset $S\subset V\setminus \{a,b\} $ with $|S|\leq 2\eta$ and $k \in \Yc^{|S|}$, the probability  matrix $ M_{(a,b)|H, \{S;k\}}:=[P(\bfY_{a,b}=i|H=j, \bfY_S=k)]_{i,j} \in \Rbb^{d^2\times r}$ has rank $r$.

\item[(A7)]{\bf Existence of an Isolated Node: }There exists a node $u_*\in V$ which is isolated from all other nodes in $G_{\cup}=\cup_{h=1}^r G_h$, i.e. \beq \label{eqn:isolated2}Y_{u_*}\indep \bfY_{V\setminus u_*} | H.\eeq

\item[(A8)]{\bf Spectral Bounds and Random Rotation Matrix: }Refer to  various spectral bounds used to obtain $K(\delta;p,d,r)$ in Appendix~\ref{proof:findcomp}, where $\delta\in (0,1)$ is fixed. Further assume that the rotation matrix $Z\in \R^{r\times r}$ in $\FindComponents$ is chosen uniformly over the Stiefel manifold $\{Q\in \R^{r\times r}: Q^\top Q =I\}$.

\item[(A9)]{\bf Number of Samples: }For  fixed $\delta,\epsilon\in (0,1)$, the number of samples satisfies \beq\label{eqn:nspect} n > n_{\spect}(\delta,\epsilon;p,d,r):=
    \frac{4K^2(\delta;p,d,r)}{\epsilon^2},\eeq
where $K(\delta;p,d,r)$ is defined in \eqref{eqn:K}.
\een

Assumption (A6) is a natural condition  required for the success of spectral decomposition, and is imposed in  ~\citep{mossel2005learning},~\citep{hsu2008spectral} and~\citep{AnandkumarHsuKakade:COLT12}. It is also known that learning singular models, i.e., those which do not satisfy (A6), is at least as hard as learning parity with noise, which is conjectured to be computationally hard~\citep{mossel2005learning}.
The condition in (A7)  is indeed an additional constraint on graph $G_{\cup}$, but is required to ensure alignment of hidden labels over   spectral decompositions of different groups of variables, as discussed before\footnote{(A7) can be relaxed as follows: if graph $G_{\cup}$ has at least three connected components, then we can choose a reference node in each of the components and estimate the marginals in the other components. For instance, if $C_1, C_2, C_3$ are three connected components in $G_{\cup}$, then we can choose a node in $C_1$ as the reference node to estimate the marginals of $C_2$ and $C_3$. Similarly, we can choose a node in $C_2$ as a reference node and estimate the marginals in $C_1$ and $C_3$.  We can then align these different estimates and obtain all the marginals.} Condition (A8) assumes various spectral bounds and (A9) characterizes the sample complexity.

\subsubsection{Guarantees for Learning Mixture Components}

We now provide the result on the success of recovering the tree approximation $T_h$ of each mixture component $P(\bfy|H=h)$. Let $\|\cdot\|_2$ on a vector denote the $\ell_2$ norm.

\bt[Guarantees for $\FindComponents$]\label{thm:spectdecomp} Under the assumptions (A1)--(A9), the procedure    in Algorithm~\ref{algo:spectral} outputs  $\hP^{\spect}(Y_a,Y_b|H=h)$, for each $a, b\in V$,    such that for all $h\in [r]$, there exists a permutation $\tau(h)\in [r]$ with \beq\|\hP^{\spect}(Y_a,Y_b|H=h)   - P(Y_a,Y_b|H=\tau(h))\|_2\leq  \epsilon,\eeq with probability at least $1-4\delta$.  \et


\bprf The proof is given in Appendix~\ref{proof:spectdecomp}.\eprf

\paragraph{Remarks: }Recall that $p$ denotes the number of variables, $r$ denotes the number of mixture components, $d$ denotes the dimension of each node variable and $\eta$ denotes the bound on separator sets between any node pair in the union graph. The quantity $K(\delta;p,d,r)$ in \eqref{eqn:K} in Appendix~\ref{proof:findcomp} is $O\left(p^{2\eta+2}d^{2\eta}r^5\delta^{-1}\poly
\log(p, d, r, \delta^{-1})\right)$. Thus, we require the number of samples scaling in \eqref{eqn:nspect} as $n = \Omega\left(p^{4\eta+4}d^{4\eta}r^{10}\delta^{-2}\epsilon^{-2}\poly
\log(p, d, r, \delta^{-1})\right)$. Since we operate in the regime where $\eta=O(1)$ is a small constant, this implies that we have a polynomial sample complexity in $p, d, r$. Note that the special case of $\eta=0$ corresponds to the case of mixture of product distributions, and it has the best sample complexity.

\subsubsection{Analysis of Tree Approximation}

We now consider the final stage of our approach, \viz learning tree approximations using  the estimates of the pairwise marginals of the mixture components from the spectral decomposition method.  We now impose a standard condition of non-degeneracy on each mixture component  to guarantee  the existence of a unique tree structure corresponding to the  maximum-likelihood tree approximation to the mixture component.

\ben \item[(A10)]{\bf Separation of Mutual Information: }Let $T_h$ denote the Chow-Liu tree corresponding to the model $P(\bfy|H=h)$ when exact statistics are input  and let
\beq \label{eqn:vartheta}\vartheta:=\min_{h\in [r]}\min_{(a,b)\notin T_h} \min_{(u,v) \in \Path(a,b;T_h)}\left(I(Y_u, Y_v|H=h)- I(Y_a,Y_b|H=h)\right),\eeq where $\Path(a,b;T_h)$ denotes the edges along the path connecting $a$ and $b$ in $T_h$.

\item[(A11)]{\bf Number of Samples: }For $\epsilon^{\tree}$ defined in \eqref{eqn:epsilontree}, the number of samples is now required to satisfy  \beq n > n_{\spect}(\delta,\epsilon^{\tree};p,d,r), \eeq where $n_{\spect}$ is given by \eqref{eqn:nspect}.

\een

The condition in (A10) assumes a  separation between mutual information along edges and non-edges of the Chow-Liu tree $T_h$ of each component model $P(\bfy|H=h)$. The quantity $\vartheta$ represents the minimum separation between the mutual information along an edge and any set of non-edges which can replace the edge in $T_h$. Note that $\vartheta\geq 0$ due to the max-weight spanning tree property of $T_h$ (under exact statistics). Intuitively $\vartheta$ denotes the ``bottleneck''  where errors are most likely to occur in tree structure estimation. Similar observations were made  by~\citet{Tan&etal:09ITsub} for error exponent analysis of Chow-Liu algorithm. The sample complexity for correctly estimating $T_h$ using samples  is based on $\vartheta_h$ and given in (A11).  This ensures that   the mutual information quantities are estimated  within the  separation bound $\vartheta$.

\bt[Tree Approximations of Mixture Components]\label{thm:treeapprox} Under  (A1)--(A11), the Chow-Liu algorithm outputs the correct   tree structures corresponding to maximum-likelihood tree approximations of the mixture components $\{P(\bfy|H=h)\}$  with probability at least $1-4\delta$, when the estimates of pairwise marginals $\{\hP^{\spect}(Y_a,Y_b|H=h)\}$ from spectral decomposition method are input.\et

\bprf See Section~\ref{sec:MI}. \eprf
\paragraph{Remarks: }Thus   our approach succeeds in recovering the correct tree structures corresponding to ML-tree approximations of mixture components with computational and sample complexities scaling  polynomially in the number of variables $p$, number of components $r$ and the dimension of each variable $d$.

Note that if the underlying model is a tree mixture, we recover the tree structures of the mixture components. For this special case, we can give a slightly better guarantee by  estimating Chow-Liu trees which are subgraphs of the union graph estimate $\hG_{\cup}$, and this is discussed in Appendix~\ref{sec:treemixture}. The improved bound $K^{\tree}(\delta;p,d,r)$ is $O\left(p^{ 2}(d\Delta)^{2\eta}r^5\delta^{-1}\poly
\log(p, d, r, \delta^{-1})\right)$, where $\Delta$ is the maximum degree in $G_{\cup}$.

\section{Conclusion}

In this paper, we considered learning tree approximations of  graphical model mixtures. We proposed novel methods which combined techniques used previously in graphical model selection, and in learning  mixtures of product distributions. We provided provable guarantees for our method, and established that it has polynomial sample and computational complexities in the number of nodes $p$, number of mixture components $r$ and cardinality of each node variable $d$. Our guarantees are applicable for a wide family of models. In future, we plan to investigate learning mixtures of continuous models, such as Gaussian mixture models.

\subsubsection*{Acknowledgements}

The first author is supported in part by the setup funds at UCI and by the AFOSR Award FA9550-10-1-0310.
\begin{appendix}
\section{Implementation of Spectral Decomposition Method}\label{sec:specdetails}

\begin{figure}[h]
\centering{
\bp\psfrag{w1}[l]{\tccyan{$a$}}\psfrag{w2}[l]{\tccyan{$b$}}
\psfrag{c}[c]{$\tcr{c}$}
\psfrag{H}[l]{\tcdkg{$H$}}\psfrag{G}[c]{$G_{\cup}$}\psfrag{u}[l]{\tcb{$u_*$}}
\psfrag{S}[c]{$\tcv{S}$}
\includegraphics[width=2in]{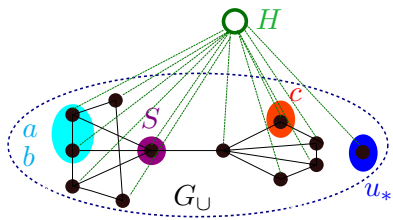}
\ep}
\caption{By conditioning on the separator set $S$ on the union graph $G_{\cup}$, we have a mixture of product distribution with respect to nodes $\{u_*,c,(a,b)\}$, i.e., $Y_{u_*}\indep Y_c\indep Y_{a,b}|\bfY_S, H$.}\label{fig:mixtureseparation-2}\end{figure}

\paragraph{Overview of the algorithm: }We provide the procedure in Algorithm~\ref{algo:spectral}. The algorithm computes the pairwise statistic of each node pair $a,b\in V\setminus \{u_*\}$, where $u_*$ is the reference node which is  isolated   in $\hG_{\cup}$, the union  graph
estimate obtained using Algorithm~\ref{algo:ranktest}. The spectral decomposition is carried out on the triplet $\{u_*, c, (a,b); \{S=k\}\}$, where   $c$ is any node not in the neighborhood of $a$ or $b$ in graph $\hG_{\cup}$. Set $S\subset V\setminus \{a, b, u_*\}$ is separates $a$, $b$ from $c$ in $\hG_{\cup}$. See
Fig.\ref{fig:mixtureseparation-2}. We fix the configuration of the separator set to $\bfY_S=k$, for each $k \in \Yc^{|S|}$, and consider the empirical distribution of $n$ samples,  $\hP^n(Y_{u_*}, Y_a, Y_a, Y_c, \{\bfY_S=k\})$. Upon spectral decomposition, we obtain the  mixture components $\hP^{\spect}(Y_a, Y_b, \bfY_S|H=h)$ for $h\in[r]$. We can then employ the estimated pairwise marginals to find the Chow-Liu tree approximation $\{\hT_h\}_h$ for each mixture component.  This routine can also be adapted to estimate the individual Markov graphs $\{G_h\}_h$ and is described briefly in Section~\ref{sec:graphest}. Also, if the underlying model is a tree mixture, we can slightly modify the algorithm and obtain better guarantees, and we outline it in Section~\ref{sec:graphest}.

\begin{algorithm}[h]\begin{algorithmic}
\caption{$\FindComponents(\bfy^n, \hG;r)$
for finding the tree-approximations of the components $\{P(\bfy|H=h)\}_h$   of  an $r$-component mixture using samples  $\bfy^n$   and  graph   $\hG$, which is an estimate of the graph $G_{\cup}:=\cup_{h=1}^r G_h$ obtained using Algorithm~\ref{algo:ranktest}.} \label{algo:spectral}
\STATE  $\hM^n_{A,B, \{C;k\}}:=[P(\bfY_A=i, \bfY_B=j, \bfY_C=k]_{i,j}$ denotes the empirical joint probability matrix estimated using samples $\bfy^n$, where $A\cap B\cap C=\emptyset$. Let $\Sc(A, B;G_{\cup})$ be a minimal vertex separator separating  $A$ and $B$ in graph $\hG_{\cup}$.
\STATE Choose a uniformly random orthonormal basis $\{z_1, \ldots, z_r\}\in \Rbb^r$. Let $Z\in \Rbb^{r\times r}$ be a matrix whose $l^{\tha}$ row is $\bfz_l^\top$.
\STATE Let $u_*\in V$ be isolated from all the other nodes in graph $\hG$. Otherwise declare fail.
\FOR{$a,b\in V\setminus \{u_*\}$}
\STATE Let $c \notin \nbd(a;\hG)\cup \nbd(b;\hG)$ (if no such node is found, go to the next node pair).
 $S\leftarrow \Sc((a,b), c;\hG)$.
\STATE $\{\hP^{\spect}(Y_a,Y_b,\bfY_S|H=h)\}_{h}\leftarrow\SpectDecomp(  u_*, c, (a,b); S,  \bfy^n,r,Z)$. 
\ENDFOR
\FOR{$h\in [r]$}
\STATE $\left[\hT_h, \{\hP^{\tree}(Y_a,Y_b|H=h)\}_{(a,b)\in \hT_h}\right] \leftarrow \TreeApprox\left(\{\hP^{\spect}(Y_a, Y_b|H=h)\}_{a,b\in V\setminus \{u_*\}}\right)$.
\ENDFOR
\STATE Output $\left[\h\pibf^{\spect}_H(h),\,\,\hT_h,\,\, \left\{\hP^{\tree}(Y_a,Y_b|H=h):(a,b)\in \hT_h\right\}\right]_{h\in [r]}$.
 \end{algorithmic}
\end{algorithm}

\floatname{algorithm}{Procedure}

\begin{algorithm}[h]\begin{algorithmic}
\caption{$[\{\hP(Y_w, \bfY_S|H=h), \h\pibf_H(h)\}_{h}]\leftarrow\SpectDecomp(u,v,w; S, \bfy^n,r, Z)$
for finding the components   of  an $r$-component mixture  from   $\bfy^n$ samples at  $w$, given witnesses $u,v$ and separator $S$ on   graph   $\hG^n$.} \label{algo:spectraldecomp}
\STATE Let $\hM^n_{u,v,\{S;k\}}:=[\hP^n(Y_u=i,Y_v=j, \bfY_S=k)]_{i,j}$ where $\hP^n$ is the empirical distribution computed using  samples $\bfy^n$. Similarly, let $\hM^n_{u,v,\{S;k\},\{w;l\}}:=[\hP^n(Y_u=i,Y_v=j,\bfY_S=k, Y_w=l)]_{i,j}$. For a vector $\lambdabf$, let $\Diag(\lambdabf)$ denote the corresponding diagonal matrix.
\FOR{$k \in \Yc^{|S|}$}
\STATE Choose  $U_{u}$ as the set of top $r$ left orthonormal singular vectors of  $\hM^n_{ u,v,\{S;k\}}$ and   $V_v$ as the right singular vectors. Similarly  for node $w$, let $U_w$ be the top $r$ left orthonormal singular vectors of $\hM^n_{w, u,\{S;k\}}$.
\FOR{$l\in [r] $}
\STATE $\bfm_l \leftarrow U_w \bfz_l$, $A\leftarrow U^\top_u\hM^n_{ u,v,\{S;k\}}V_v$ and $B_l\leftarrow U^\top_u\left( \sum_{q} m_l(q)\hM^n_{ u,v,\{S;k\},\{w;q\}}\right)V_v$.
\IF{$A$  is invertible (Fail Otherwise)}
\STATE $C_l \leftarrow B_l A^{-1}$.
$\Diag(\lambdabf^{(l)}) \leftarrow R^{-1} C_l R$. \COMMENT{Find  $R$ which diagonalizes $C_l$ for the first triplet. Use the same matrix $R$ for all other triplets.}
\ENDIF
\ENDFOR
\STATE Form the matrix from the above eigenvalue computations: $\Lambda=[\lambdabf^{(1)}|\lambdabf^{(2)}|\ldots, \lambdabf^{(r)}]$
\STATE Obtain $\hM_{w|H, \{S;k\}}\leftarrow U_w Z^{-1} \Lambda^\top$. Similarly obtain $\hM_{v|H,\{S;k\}}$.
\STATE Obtain   $\h\pibf_H$: $\hM^n_{v,w,\{S;k\}}= \hM_{v|H, \{S;k\}} \Diag(\h\pibf_{H|\{S;k\}}) (\hM_{w|H,\{S;k\}})^\top \hP^n(\bfY_S=k)$.
 \ENDFOR
\STATE Output $\{\hP(Y_w,\bfY_S|H=h), \h\pibf_H(h)\}_{h\in[ r]}$.
 \end{algorithmic}
\end{algorithm}

\begin{algorithm}[h]\begin{algorithmic}
\caption{$[\hT, \{\hP^{\tree}(Y_a, Y_b)\}_{(a,b)\in \hT}]\leftarrow\TreeApprox(\{\hP(Y_a, Y_b)\}_{a,b\in V\setminus \{u_*\}}$ for finding a tree approximation given the pairwise statistics.}\label{algo:cl}
\FOR{$a,b\in V\setminus \{u_*\}$}
\STATE Compute mutual information $\hI(Y_a;Y_b)$ using $\hP(Y_a, Y_b)$.
\ENDFOR
\STATE $\hT\leftarrow \MaxWtTree(\{\hI(Y_a;Y_b)\})$ is  max-weight spanning tree using edge weights $\{\hI(Y_a;Y_b)\}$.
\FOR{$(a,b)\in \hT$}
\STATE $\hP^{\tree}(Y_a, Y_b)\leftarrow \hP(Y_a, Y_b)$.
\ENDFOR
 \end{algorithmic}
\end{algorithm}


\subsection{Discussion and Extensions}\label{sec:graphest}
\paragraph{Simplification for Tree Mixtures $(G_h=T_h)$: }We can simplify the above method by limiting to tree approximations which are subgraphs of graph $G_{\cup}$. This procedure coincides with the original method  when all the component Markov graphs $\{G_h\}_h$ are trees, i.e., $G_h=T_h$, $h\in [r]$. This is because in this case,  the Chow-Liu tree coincides with $T_h\subset G_{\cup}$ (under exact statistics). This implies that we need to compute pairwise marginals {\em only} over the edges of $G_{\cup}$ using $\SpectDecomp$ routine, instead of over all the node pairs, and the $\TreeApprox$ procedure computes a maximum weighted spanning tree over $G_{\cup}$, instead of the complete graph. This leads a slight improvement of sample complexity, and we note it in the remarks after Theorem~\ref{thm:spectdecomp}.

 \paragraph{Estimation of Component Markov Graphs $\{G_h\}_h$: }
We now note that we can also estimate the component Markov graphs $\{G_h\}$ using the spectral decomposition routines and we briefly describe it below. Roughly, we can do a suitable conditional independence test on the estimated statistics $\hP^{\spect}(\bfY_{\nbd[a;\hG_{\cup}]}|H=h)$ obtained from spectral decomposition,  for each node neighborhood $\nbd[a;\hG_{\cup}]$, where $a\in V\setminus \{u_*\}$ and $\hG_{\cup}$ is an estimate of $G_{\cup}:=\cup_{h\in [r]} G_h$. We can estimate these statistics by selecting a suitable set of witnesses $C:=\{c_1,c_2, \ldots, \}$ such that $\nbd[a]$ can be separated from $C$ in  $\hG_{\cup}$. We can employ Procedure $\SpectDecomp$ on this configuration by using a suitable separator set and  then doing a threshold test on the estimated component statistics $\hP^{\spect}$: if   for each $(a,b)\in \hG_{\cup}$, the following quantity\[ \min_{\substack{k,l\in \Yc}} \norm{\hP^{\spect}(Y_a|Y_b=k,\bfY_{\nbd(a)\setminus b}=\bfy, H=h) - \hP^{\spect}(Y_a|Y_b=l, \bfY_{\nbd(a)\setminus b}=\bfy, H=h)}_1,\]  is below a certain threshold, for some $\bfy\in \Yc^{|\nbd(a)\setminus b|}$, then it is  removed  from $\hG_{\cup}$, and we obtain  $\hG_h$ in this manner.  A similar test was used for graphical model selection (i.e., not a mixture model) in~\citep{Bresler&etal:Rand}. We note that we can obtain sample complexity results for the above test, on lines of the analysis in Section~\ref{sec:exactsepspectral} and this method is efficient when the maximum degree in $G_{\cup}$ is small.

\section{Extension to Graphs with Sparse Local Separators}\label{sec:corrdecay}

\subsection{Graphs with Sparse Local Separators}

We now extend the analysis to the setting where the graphical model mixture has the union  graph $G_{\cup}$ with sparse local separators, which is a weaker criterion than having sparse exact separators. We now provide the definition of a local separator. For detailed discussion, refer to~\citep{AnandkumarTanWillsky:Ising11}.

For $\gamma\in\Nbb$, let $B_\gamma(i;G )$ denote the set of vertices within distance $\gamma$ from $i$ with respect to graph $G$. Let  $F_{\gamma, i}:=G(B_\gamma(i))$ denote the subgraph of $G$ spanned by $B_\gamma(i;G)$, but in addition, we retain the nodes not in $B_\gamma(i)$ (and remove the corresponding edges).

\bd[$\gamma$-Local Separator]\label{def:localsep} Given a graph $G$, a $\gamma$-{\em local separator} $\Sc_{\local}(i,j;G, \gamma)$ between $i$ and $j$,   for $(i,j)\notin G$,  is a {\em minimal} vertex separator\footnote{A minimal separator is a separator of smallest cardinality.} with respect to the subgraph $F_{\gamma,i} $. In addition,  the parameter $\gamma$ is referred to as the {\em path threshold} for  local separation. A graph is said to be $\eta$-locally separable, if
\beq \max_{(i,j)\notin G}| \Sc_{\local}(i,j;G,\gamma)|\leq \eta.\eeq
\ed

A wide family of graphs possess the above property of sparse local separation, i.e., have a small $\eta$. In addition to graphs considered in the previous section, this additionally  includes the family of locally tree-like graphs (including sparse random graphs), bounded degree graphs, and augmented graphs, formed by the union of a bounded degree graph and a locally tree-like graph (e.g. small-world graphs). For detailed discussion, refer to~\citep{AnandkumarTanWillsky:Ising11}.

\subsection{Regime of Correlation Decay}

We consider learning mixtures of graphical models Markov on graphs with sparse local separators. We assume that these models are in the regime of correlation decay, which makes learning feasible via our proposed methods. Technically, correlation decay can be defined in multiple ways and we use the notion of {\em strong spatial mixing}~\citep{Weitz:STOC06}. A weaker notion is known as {\em weak spatial mixing}.

A graphical model is said to satisfy   weak spatial mixing when the   conditional distribution at each  node $v$ is asymptotically independent of the configuration of a growing  boundary (with respect to $v$). It is said to satisfy strong spatial mixing, when the total variation distance   between   two conditional distributions at each node $v$, due to conditioning on different configurations,    depends {\em only} on the graph distance  between node $v$ and the set  where the two configurations differ. We formally define it below\footnote{We slightly modify the definition of correlation decay   compared to the usual notion by considering models on different graphs, where one is an induced subgraph of the neighborhood  of the other graph, instead of models with different boundary conditions.} and incorporate it to provide learning guarantees. See~\citep{Weitz:STOC06} for details.

Let $P(Y_v|\bfY_A;G)$ denote the conditional distribution of node $v$ given a set $A\subset V\setminus \{v\}$ under model $P$ with Markov graph $G$.  For some subgraph $F\subset G$, let $P(Y_v|\bfY_A;F)$ denote the   conditional  distribution on  corresponding to a graphical model Markov on subgraph $F$ instead of $G$, i.e., by setting the  potentials of edges (and hyperedges)  in $G\setminus F$ to zero. For any two sets $A_1, A_2\subset V$, let $\dist(A_1, A_2):=\min_{u\in A_1, v \in A_2} \dist(u,v)$ denote the minimum graph distance. Let $B_l(v)$ denote the set of nodes within graph distance $l$ from node $v$ and $\partial B_l(v)$ denote the boundary nodes, i.e., exactly at $l$ from node $v$.
Let $F_l(v;G):=G(B_l(v))$ denote the induced subgraph on $B_l(v;G)$.
For any  vectors $\bfa,\bfb$, let $\norm{\bfa-\bfb}_1:= \sqrt{\sum_i |a(i)-b(i)|}$ denote the $\ell_1$ distance between them.

\bd[Correlation Decay]\label{def:corrdecay}A graphical model $P$ Markov on graph $G=(V, E)$ with $p$ nodes is said to exhibit correlation decay with a non-increasing rate function   $\zeta(\cdot)>0$ if for all $l,p \in \Nbb$, \beq  \max_{\substack{v \in V\\ A\subset V\setminus \{v\}} }\norm{P(Y_v|\bfY_A=\bfy_A;G)-
P(Y_V|\bfY_{A}=\bfy_A;F_l(i;G))}_1 = \zeta(\dist(A,\partial B_l(i))).\label{eqn:corrdecay}\eeq\ed

\vspace{1em}

\noindent{\bf Remarks: }\ben


\item In \eqref{eqn:corrdecay}, if we consider the marginal distribution of node $v$ instead of its conditional distribution over all sets $A$, then we have a weaker criterion, typically referred to as {\em weak spatial mixing}. However, in order to provide learning guarantees, we require the notion of strong mixing.


\item For the  class of Ising models (binary variables), the regime of correlation decay can be explicitly characterized, in terms of the maximum edge potential of the model. When the maximum edge potential is below a certain threshold, the model is said to be in the regime of correlation decay. The threshold that can be explicitly characterized for certain graph families. See~\citep{AnandkumarTanWillsky:Ising11} for derivations.

\een

\subsection{Rank Test Under Local Separation}\label{sec:assumptions-corrdecay}

We now provide sufficient conditions for the success of $\ranktest(\bfy^n;\xi_{n,p}, \eta, r)$ in Algorithm~\ref{algo:ranktest}. Note that the crucial difference compared to the previous section is that $\eta$ refers to the bound on local separators in contrast to the bound on exact separators. This can lead to significant reduction in computational complexity of running the rank test for many graph families, since the complexity scales as $O(p^{\eta+2} d^3)$ where $p$ is the number of nodes and $d$ is the cardinality of each node variable.

\ben

\item[(B1)]{\bf Number of Mixture Components: } The number of components $r$ of the mixture model and dimension $d$ of each node variable  satisfy \beq d> r.\eeq The mixing weights of the latent factor $H$ are assumed to be strictly positive\[\pi_H(h):= P(H=h)>0, \quad \forall\,\, h\in[ r].\]

\item[(B2)]{\bf Constraints on Graph Structure: }Recall that  $G_{\cup}=\cup_{h=1}^r G_h$ denotes the union of the Markov graphs of the mixture components and we assume that $G_{\cup}$ is $\eta$-locally separable according to Definition~\ref{def:localsep}, i.e.,  for the chosen path threshold $\gamma\in \Nbb$, we assume that\[ |\Sc_{\local}(u,v;G_{\cup},\gamma)|\leq \eta=O(1), \quad \forall (u,v)\notin G_{\cup}.\]

\item[(B3)]{\bf Rank Condition: }We assume that the matrix $M_{u,v,\{S;k\}}$ in \eqref{eqn:M} has rank strictly greater than $r$ when the nodes $u$ and $v$  are neighbors in graph $G_{\cup}=\cup_{h=1}^r G_h$ and the set satisfies $|S|\leq \eta$. Let $\rho_{\min}$ denote  \beq \label{eqn:rhominlocal}\rho_{\min}:= \min_{\substack{(u,v)\in G_{\cup}, |S|\leq \eta \\ S\subset V\setminus \{u,v\}}} \max_{k\in \Yc^{|S|}}\sigma_{r+1}\left(M_{u,v,\{S;k\}}\right)>0.\eeq

\item[(B4)]{\bf Regime of Correlation Decay: }We assume that all the mixture components $\{P(\bfy|H=h;G_h)\}_{h\in [r]}$ are in the regime of correlation decay according to  Definition~\ref{def:corrdecay} with rate functions $\{\zeta_{h}(\cdot)\}_{h\in [r]}$. Let \beq \label{eqn:zetatot} \zeta(\gamma):= 2\sqrt{d}\max_{h \in [r]}  \zeta_{h}(\gamma).\eeq We assume that the minimum singular value $\rho_{\min}$ in \eqref{eqn:rhomin} and $\zeta(\gamma)$ above satisfy $\rho_{\min}> \zeta(\gamma)$.

\item[(B5)] {\bf Choice of threshold $\xi$: }For $\ranktest$ in Algorithm~\ref{algo:ranktest}, the threshold $\xi$ is chosen as \[ \xi := \frac{\rho_{\min}- \zeta(\gamma)}{2}>0, \] where $\zeta(\gamma)$ is given by \eqref{eqn:zetatot}  and $\rho_{\min}$ is given by  \eqref{eqn:rhomin}, and $\gamma$ is the path threshold for local separation on graph $G_{\cup}$.

\item[(B6)] {\bf Number of Samples: }Given an $\delta>0$, the number of samples $n$ satisfies \beq n > n_{\lrank}(\delta;p):=  \max\left( \frac{1}{t^2}\left(2 \log p + \log \delta^{-1} +\log 2\right),   \left(\frac{2}{\rho_{\min}-  \zeta(\gamma)-t}\right)^2\right), \eeq where $p$ is the number of nodes, for some    $t\in (0, \rho_{\min}  - \zeta(\gamma))$.

\een

The above assumptions (B1)--(B6) are comparable to assumptions (A1)--(A5) in Section~\ref{sec:assumptions-rank-exact}. The conditions on $r$ and $d$ in (A1) and (B1) are identical. The conditions (A2) and (B2) are comparable, with the only difference being that (A2) assumes bound on exact separators while (B2) assumes bound on local separators, which is a weaker criterion. Again, the conditions (A3) and (B3) on the rank of matrices for neighboring nodes are identical. The condition (B4) is an additional condition regarding the presence of correlation decay in the mixture components. This assumption is required for approximate conditional independence under conditioning with local separator sets in each mixture component. In addition, we require that $\zeta(\gamma) < \rho_{\min}$. In other words, the threshold $\gamma$ on path lengths considered for local separation should be large enough (so that the corresponding value $\zeta(\gamma)$ is small). (B5) provides a modified threshold to account for distortion due to the use of local separators and (B6) provides the modified sample complexity.

\subsubsection{Success of Rank Tests}

We now provide the result on the success of recovering the union graph $G_{\cup}:=\cup_{h=1}^r G_h$ for $\eta$-locally separable graphs.

\bt[Success of Rank Tests]\label{thm:ranktestapprox}  The $\ranktest(\bfy^n;\xi,\eta, r)$ outputs the correct graph $G_{\cup}:=\cup_{h=1}^r G_h$, which is the union of the component Markov graphs, under the assumptions (B1)--(B6) with probability at least $1-\delta$.\et

\bprf See Appendix~\ref{proof:ranktest}. \eprf

\subsection{Results for Spectral Decomposition Under Local Separation}\label{sec:approxsepspectral}

The $\FindComponents(\bfy^n, \hG;r)$ procedure in Algorithm~\ref{algo:spectral} can also be implemented for graphs with local separators, but with the modification that we use local separators $\Sc_{\local}((a,b),c;\hG)$, as opposed to exact separators, between nodes $a,b$ and $c$ under consideration. We prove that this method succeeds in estimating the pairwise marginals of the component model under the following set of conditions. We find that there is additional distortion introduced due to the use of local separators in $\FindComponents$ as opposed to exact separators.

\subsubsection{Assumptions}
In addition to the assumptions (B1)--(B6), we impose the following constraints to guarantee the success of estimating the various   mixture components.

\ben

\item[(B7)]{\bf Full Rank Views of the Latent Factor: }For each node pair $a,b\in V$, and any subset $S\subset V\setminus \{a,b\} $ with $|S|\leq 2\eta$ and $k \in \Yc^{|S|}$, the probability  matrix $ M_{(a,b)|H, \{S;k\}}:=[P(\bfY_{a,b}=i|H=j, \bfY_S=k)]_{i,j} \in \Rbb^{d^2\times r}$ has rank $r$.

\item[(B8)]{\bf Existence of an Isolated Node: }There exists a node $u_*\in V$ which is isolated from all other nodes in $G_{\cup}=\cup_{h=1}^r G_h$, i.e. \beq \label{eqn:isolatedlocal}Y_{u_*}\indep \bfY_{V\setminus u_*} | H.\eeq

\item[(B9)]{\bf Spectral Bounds and Random Rotation Matrix: }Refer to  various spectral bounds used to obtain $K(\delta;p,d,r)$ in Appendix~\ref{proof:findcomp}, where $\delta\in (0,1)$ is fixed. Further assume that the rotation matrix $Z\in \R^{r\times r}$ in $\FindComponents$ is chosen uniformly over the Stiefel manifold $\{Q\in \R^{r\times r}: Q^\top Q =I\}$.

\item[(B10)]{\bf Number of Samples: }For  fixed $\delta\in (0,1)$ and $\epsilon >\epsilon_0$ , the number of samples satisfies \beq\label{eqn:nlspect} n > n_{\lspect}(\delta,\epsilon;p,d,r):=
    \frac{4K^2(\delta;p,d,r)}{\left(\epsilon-\epsilon_0\right)^2},\eeq
where \beq \label{eqn:epsilon0}\epsilon_0 := 2K'(\delta;p,d,r)\zeta(\gamma),\eeq and  $K'(\delta;p,d,r)$ and $K(\delta;p,d,r)$  are defined in \eqref{eqn:K'} and \eqref{eqn:K}, and $\zeta(\gamma)$ is given by \eqref{eqn:zetatot}.
\een

The assumptions (B7)-(B9) are identical with (A6)-(A8). In (B10), the bound on the number of samples is slightly worse compared to (A9), depending on the correlation decay rate function $\zeta(\gamma)$. Moreover, the perturbation $\epsilon$ now has a lower bound $\epsilon_0$ in \eqref{eqn:epsilon0}, due to the use of local separators in contrast to exact vertex separators.
As before, below, we impose additional conditions in order to obtain the correct Chow-Liu tree approximation $T_h$ of each mixture component $P(\bfy|H=h)$.

\ben \item[(B11)]{\bf Separation of Mutual Information: }Let $T_h$ denote the Chow-Liu tree corresponding to the model $P(\bfy|H=h)$ when exact statistics are input\footnote{Assume that the Chow-Liu tree $T_h$ is unique for each component $h \in [r]$ under exact statistics, and this  holds for generic parameters.} and let
\beq \label{eqn:varthetalocal}\vartheta:=\min_{h\in [r]}\min_{(a,b)\notin T_h} \min_{(u,v) \in \Path(a,b;T_h)}\left(I(Y_u, Y_v|H=h)- I(Y_a,Y_b|H=h)\right),\eeq where $\Path(a,b;T_h)$ denotes the edges along the path connecting $a$ and $b$ in $T_h$.

\item[(B12)]{\bf Constraint on Distortion: }
 For function $\phi(\cdot)$ defined  in \eqref{eqn:phi} in Appendix~\ref{sec:MI}, and  for some $\tau\in (0, 0.5\vartheta)$, let $ \epsilon^{\tree}:=\phi^{-1}\left(\frac{ 0.5 \vartheta -\tau}{3 d}\right)>\epsilon_0,$ where $\epsilon_0$ is given by \eqref{eqn:epsilon0}.  The number of samples is now required to satisfy  \beq n > n_{\lspect}(\delta,\epsilon^{\tree};p,d,r), \eeq where $n_{\lspect}$ is given by \eqref{eqn:nlspect}.

\een

Conditions (B11) and (B12) are identical to (A10) and (A11), except that the required bound $\epsilon^{\tree}$ in (B12) is required to be above the lower bound $\epsilon_0$ in \eqref{eqn:epsilon0}.

\subsubsection{Guarantees for Learning Mixture Components}

We now provide the result on the success of recovering the tree approximation $T_h$ of each mixture component $P(\bfy|H=h)$ under local separation.

\bt[Guarantees for $\FindComponents$]\label{thm:spectdecomplocal} Under the assumptions (B1)--(B10), the procedure    in Algorithm~\ref{algo:spectral} outputs  $\hP^{\spect}(Y_a,Y_b|H=h)$, for $a, b\in V\setminus \{u_*\}$,  with probability at least $1-4\delta$, such that for all $h\in [r]$, there exists a permutation $\tau(h)\in [r]$ with \beq\|\hP^{\spect}(Y_a,Y_b|H=h)   - P(Y_a,Y_b|H=\tau(h))\|_2\leq  \epsilon.\eeq Moreover, under additional assumptions (B11)-(B12), the method outputs the correct Chow-Liu tree $T_h$ of each component $P(\bfy|H=h)$ with probability at least $1-4\delta$.\et

\vspace{1em}

\noindent{\bf Remark: }The sample and computational complexities are significantly improved, since it only depends on  the size of local separators (while previously it depended on the size of exact separators).


\section{Analysis of Rank Test: Proof of Theorem~\ref{thm:ranktest} and~\ref{thm:ranktestapprox}}\label{proof:ranktest}
\paragraph{Bounds on Empirical Probability: }
We first recap the result from~\citep[Proposition 19]{hsu2008spectral}, which is an application of the McDiarmid's inequality. Let $\norm{\cdot}_2$ the $\ell_2$ norm of a vector.

\begin{proposition}[Bound for Empirical Probability Estimates]\label{prop:mcdiarmid}Given empirical estimates $\hP^n$ of a probability vector $P$ using $n$ i.i.d.\ samples, we have \beq\Pbb[\norm{\hP^n-P}_2>\epsilon]\leq \exp\left[-n\left(\epsilon-1/\sqrt{n}\right)^2\right],\quad\forall\, \epsilon>1/\sqrt{n}.  \eeq \end{proposition}

\vspace{1em}

\noindent{\bf Remark: }The bound is independent of the cardinality of the sample space.

This implies concentration bounds for $\hM_{u,v,\{S;k\}}$. Let $\|\cdot\|_2$ and $\|\cdot\|_{\Fbb}$ denote the spectral norm and the Frobenius norms respectively.

\begin{lemma}[Bounds for $\hM_{u,v,\{S;k\}}$]\label{lemma:spectconc}Given $n$ i.i.d.\ samples $\bfy^n$, the empirical estimate $\hM^n_{u,v,\{S;k\}}:=[\hP^n[Y_u=i, Y_v=j, \bfY_S=k]]_{i,j}$ satisfies\beq \Pbb\left[\max_{\substack{l\in [ d]\\k \in \Yc^{|S|}}}|\sigma_l(\hM^n_{u,v,\{S;k\}})-\sigma_l(M_{u,v,\{S;k\}})|>\epsilon\right] \leq \exp\left[-n\left(\epsilon-1/\sqrt{n}\right)^2\right],\quad \forall\, \epsilon>1/\sqrt{n}. \eeq
\end{lemma}

\bprf Using proposition~\ref{prop:mcdiarmid}, we have
\beq\Pbb[\max_{k \in \Yc^{|S|}}\norm{\hP^n(Y_u, Y_v, \bfY_S=k)-P(Y_u, Y_v, \bfY_S=k)}_2>\epsilon]\leq \exp\left[-n\left(\epsilon-1/\sqrt{n}\right)^2\right],\quad \epsilon>1/\sqrt{n}.  \eeq In other words, \beq\Pbb[\max_{k \in \Yc^{|S|}}\norm{\hM^n_{u,v,\{S;k\}}-M_{u,v,\{S;k\}}}_{\Fbb}>\epsilon]\leq \exp\left[-n\left(\epsilon-1/\sqrt{n}\right)^2\right],\quad \epsilon>1/\sqrt{n}.  \eeq Since $\norm{A}_2\leq \norm{A}_{\Fbb}$ for any matrix $A$ and applying the Weyl's theorem, we have the result.\eprf

From Lemma~\ref{lemma:sep} and Lemma~\ref{lemma:spectconc}, it is easy to see that
\[\Pbb[\hG_{\cup}^n \neq G_{\cup}] \leq 2 p^2 \exp\left[-n \left(\rho_{\min}/2-1/\sqrt{n}\right)^2\right],\] and we have the result. Similarly, we have Theorem~\ref{thm:ranktestapprox} from Lemma~\ref{lemma:localsep}
and Lemma~\ref{lemma:spectconc}.\qed

\section{Analysis of Spectral Decomposition: Proof of Theorem~\ref{thm:spectdecomp}}\label{proof:spectdecomp}

\subsection{Analysis Under Exact Statistics}\label{proof:spectexact}
We now prove the success of $\FindComponents$ under exact statistics. We first consider three   sets $A_1, A_2 , A_3\subset V$ such that $\nbd[A_i;G_{\cup}]\cap \nbd[A_j;G_{\cup}]=\emptyset$ for $i,j\in [3]$ and $G_{\cup}:=\cup_{h\in [r]} G_h$ is the union of the Markov graphs. Let $S\subset V\setminus \cup_i A_i$ be a multiway separator set for $A_1, A_2, A_3$ in graph $G_{\cup}$.  For $A_i$, $i \in \{1,2,3\}$, let $U_i \in \R^{d^{|A_i|} \times r}$ be a matrix such that
$U_i^\t M_{A_i|H,\{S;k\}}$ is invertible, for a fixed $k \in \Yc^{|S|}$.
Then $U_1^\t M_{A_1,A_2,\{S;k\}} U_2$ is invertible, and for all $\bfm\in \R^{d^{|A_3|}}$, the observable
  operator  $\tilC(\bfm) \in \R^{r \times r}$, given by
\beq \label{eqn:tilC1}\tilC(\bfm):= \left(U_1^\t \left(\sum_{q}m (q) M_{A_1,A_2,\{S;k\},\{A_3;q\}}\right) U_2\right) \left(U_1^\t M_{A_1, A_2, \{S;k\}}
U_2\right)^{-1}.\eeq Note that the above operator is computed in $\SpectDecomp$ procedure.
We now provide a generalization of the result in~\citep{AnandkumarHsuKakade:multiview12}.

\begin{lemma}[Observable Operator] \label{lemma:decompexact}
Under assumption (A6), the observable operator in \eqref{eqn:tilC1}
 satisfies  \beq\tilC(\bfm) = \left(U_1^T M_{A_1|H, \{S;k\}}\right)\Diag\left(M^\top_{A_3|H,\{S;k\}} \bfm\right)\left(U_1^T M_{A_1|H, \{S;k\}}\right)^{-1}.\eeq
In particular, the $r$ roots of the polynomial $\lambda \mapsto
\det(\tilC(\bfm) - \lambda I)$ are
$\{ \dotp{\bfm,M_{A_3|H, \{S;k\} }\bfe_j} : j \in [r] \}$.
\end{lemma}

\bprf
We have \[U_1^\t M_{A_1,A_2,\{S;k\}} U_2 = (U_1^\t M_{A_1|H,\{S;k\}}) \Diag(\pibf_{H,\{S;k\}}) (M_{A_2|H,\{S;k\}}^\t U_2)  \]on lines of \eqref{eqn:uv}, which is invertible by the assumptions
on $U_1$, $U_2$ and Assumption (A6). Similarly, \[U_1^\t M_{A_1,A_2,\{S;k\}, \{A_3;q\}} U_2 = (U_1^\t M_{A_1|H,\{S;k\}}) \Diag(\pibf_{H,\{S;k\},\{A_3;q\}}) (M_{A_2|H,\{S;k\}}^\t U_2),\]  and we have the result.
\eprf

The above result implies that we can recover the matrix $M_{A|H,\{S;k\}}$ for any set $A\subset V$, by using a suitable reference node, a witness and a separator set. We set the isolated node $u_*$ as the reference node (set $A_1$ in the above result).  Since we focus on recovering the edge marginals of the mixture components, we consider  each node pair $a,b\in V\setminus \{u_*\}$ (set $A_3$ in the above result),  and any node $c \notin \nbd(a;G_{\cup})\cup \nbd(b;G_{\cup})$ (set $A_2$ in the above result),  where $G_{\cup}:=\cup_{h\in [r]} G_h$, as described in $\FindComponents$. Thus, we are able to recover $M_{a,b|H,\{S;k\}}$ under exact statistics. Since $\bfY_S$ are observed, we have the knowledge of $P(\bfY_S=k)$, and can thus recover $M_{a,b|H}$ as desired. The spectral decompositions of different groups are aligned since we use the same node $u_*$, and since $u_*$ is isolated in $G_{\cup}$, fixing the variables $\bfY_S=k$ has no effect on the conditional distribution of $Y_{u_*}$, i.e., $P(Y_{u_*}|H, \bfY_S=k) =P(Y_{u_*}|H)$. Since we recover the edge marginals $M_{a,b|H}$ correctly we can recover the correct tree approximation $T_h$, for $h\in [r]$.

\subsection{Analysis of $\SpectDecomp(u,v,w;S)$ }

We first consider the success of Procedure $\SpectDecomp(u,v,w;S)$ for estimating the statistics of $w$ using node $u\in V$ as the reference node (which is conditionally independent of all other nodes given $H$) and witness $v\in V$   and separator set $S$. We will use this to provide sample complexity results on $\FindComponents$ using union bounds. The proof borrows heavily from~\citep{AnandkumarHsuKakade:multiview12}.

Recall that   $\hU_{u}$  is the set of top $r$ left orthonormal singular vectors of  $\hM^n_{ u,v,\{S;k\}}$ and   $\hV_v$ as the right orthonormal vectors.
For $l \in [r]$, let $\bfm_l=\hU_w \bfz_l$, where $\bfz_l$ is uniformly distributed in $\sphere^{r-1}$ and $\hU_w$ is the top $r$ left singular vectors of  $\hM^n_{w, u,\{S;k\}}$.   By Lemma~\ref{lemma:matrix-perturb}, we have that $ U^\top_u M_{ u,v,\{S;k\}}V_v$ is invertible. Recall the definition of  the observable operator in \eqref{eqn:tilC1}\beq \tilC_l:= \tilC(\bfm_l)= \hU^\top_u\left( \sum_{q} m_l(q) M_{ u,v,\{S;k\},\{w;q\}}\right)\hV_v \left( U^\top_u M_{ u,v,\{S;k\}}V_v\right)^{-1} ,\label{eqn:tilC}\eeq where exact matrices $M$ are used. Denote $\hC_l$ when the sample versions $\hM^n$ are used \beq \hC_l:= \hU^\top_u\left( \sum_{q} m_l(q) \hM^n_{ u,v,\{S;k\},\{w;q\}}\right)\hV_v \left( U^\top_u \hM^n_{ u,v,\{S;k\}}V_v\right)^{-1} ,\label{eqn:hC}\eeq  We have the following result.

\begin{lemma}[Bounds for $\norm{\hC_l- \tilC_l}_2$]  The matrices $\tilC_l$ and $\hC_l$ defined in \eqref{eqn:tilC} and \eqref{eqn:hC} satisfy\begin{align}\nn \norm{\hC_l- \tilC_l}_2 \leq& \frac{ 2 \norm{\sum_{q} m_l(q)(\hM^n_{u,v,\{S;k\},\{w;q\}}- M_{u,v,\{S;k\},\{w;q\}})}_2}{\sigma_r(M_{u,v,\{S;k\}})}\\ & + \frac{2 \norm{\sum_{q} m_l(q)M_{u,v,\{S;k\},\{w;q\}}}_2
\norm{\hM^n_{u,v,\{S;k\}}- M_{u,v,\{S;k\}} }_2}{\sigma_r(M_{u,v,\{S;k\}})^2}.\end{align}
\end{lemma}

\bprf Using Lemma~\ref{lemma:op-perturb} and  Lemma~\ref{lemma:decompexact}.  \eprf

We now provide perturbation bounds between estimated matrix $\hM_{w|H, \{S;k\}}$ and the true matrix $M_{w|H,\{S;k\}}$. Define
\begin{align}\label{eqn:beta}
\beta(w) &:=\min_{k\in \Yc^{|S|}}
\min_{i \in [r]}
\min_{j \neq j'}
|\dotp{\bfz^{(i)},\h{U}_w^\t M_{w|H,\{S;k\}}(\e_j-\e_{j'})}|
\\
\lambda_{\max}(w) &:=
\max_{i,j \in [r]}
|\dotp{\bfz^{(i)},\h{U}_w^\t  M_{w|H,\{S;k\}}\e_j}|
,\label{eqn:lambdamax}
\end{align}where  $\bfz_l$ is uniformly distributed in $\sphere^{r-1}$.

\begin{lemma}[Relating $\hM_{w|H, \{S;k\}}$ and   $M_{w|H,\{S;k\}}$] The estimated matrix $\hM_{w|H, \{S;k\}}$  using samples and the true matrix $M_{w|H, \{S;k\}}$  satisfy, for all $j \in [r]$, \begin{align}\nn
&\|\hM_{w|H,\{S;k\}} \bfe_j - M_{w|H,\{S;k\}} \bfe_{\tau(j)}\|_2
\leq  2\| M_{w|H,\{S;k\}} \bfe_{\tau(j)}\|_2 \cdot \frac{\|\hM^n_{u,w,\{S;k\}} -
M_{u,w,\{S;k\}}\|_2}{\sigma_r(M_{u,w,\{S;k\}})}
\\  &+
\Bigl( 12\sqrt{r} \cdot \kappa(M_{u|H})^2 +
256r^2 \cdot \kappa(M_{u|H})^4 \cdot \lambda_{\max}(w)/\beta(w) \Bigr)
\cdot \|\hC_l -\tilC_l\|_2.
\end{align}
\end{lemma}

\bprf
Define a matrix $R:= \hU_u^\top M_{u|H}\Diag(\norm{\hU_u^\top M_{u|H} \bfe_1}_2,\ldots, \norm{\hU_u^\top M_{u|H} \bfe_r}_2 )^{-1} $. Note that $R$ has unit norm columns and  $R$ diagonalizes $\tilC_l$, i.e., \[ R^{-1} \tilC_l R =\Diag(M^\top_{w|H,\{S;k\}} \bfz_l).\] Using the fact that for any stochastic matrix $d\times r$ matrix $A$, $\norm{A}_2\leq \sqrt{r} \norm{A}_1 = \sqrt{r}$ and Lemma~\ref{lemma:normalize-eig}, we have
\[\norm{R^{-1}}_2 \leq 2 \kappa(\hU_u^\top M_{u|H}), \quad \kappa(R) \leq 4 \kappa(M_{u|H}). \]
From above and by Lemma~\ref{lemma:eig-perturb-all}, there exist  a permutation $\tau$ on $[r]$ such that, for
all $j,l \in [r]$,
\begin{align}
|\h\lambda^{(l)}(j) - \lambda^{(l)}(\tau(j))|
& \leq \Bigl( 3\kappa(R) + 16r^{1.5} \cdot \kappa(R) \cdot \|R^{-1}\|_2^2
\cdot \lambda_{\max}(w)/\beta(w) \Bigr)
\cdot \|\hC_l -\tilC_l\|_2
\nonumber
\\
& \leq \Bigl( 12\kappa(M_{u|H})^2 +
256 r^{1.5} \cdot \kappa(M_{u|H})^4 \cdot \lambda_{\max}(w)/\beta(w) \Bigr)
\cdot \|\hC_l -\tilC_l\|_2,
\label{eq:lambda-error}
\end{align} where $\beta(w)$ and $\lambda_{\max}(w)$ are given by \eqref{eqn:beta} and \eqref{eqn:lambdamax}.
Let $\h\nubf^{(j)} := (\h\lambda^{(1)}(j), \h\lambda^{(2)}(j), \dotsc, \h\lambda^{(r)}(j))
\in \R^r$ be the row vector corresponding to $j^{\tha}$ row of $\hLambda$ and $\v\nu^{(j)} := (\lambda^{(1)}(j), \lambda^{(2)}(j), \dotsc,
\lambda^{(r)}(j)) \in \R^r$.
Observe that $\v\nu^{(j)} = Z \hU_{w|H,\{S;k\}}^\t M_{w|H,\{S;k\}}\e_j$.
By the orthogonality of $Z$, the fact $\|\v{v}\|_2 \leq \sqrt{r}
\|\v{v}\|_\infty$ for $\v{v} \in \R^r$, and the above inequality,
\begin{align}
\nn&\|Z^{-1}\h\nubf^{(j)} -\hU_{w|H,\{S;k\}}^\t M_{w|H,\{S;k\}}\e_{\tau(j)}\|_2
\\& = \|Z^{-1}(\h\nubf^{(j)} - \v\nu^{(\tau(j))})\|_2
\nonumber \\
& = \|\h\nubf^{(j)} - \v\nu^{(\tau(j))}\|_2
\nonumber \\
& \leq \sqrt{r} \cdot \|\h\nubf^{(j)} - \v\nu^{(\tau(j))}\|_\infty
\nonumber \\
& \leq
\Bigl( 12\sqrt{r} \cdot \kappa(M_{u|H})^2 +
256r^2 \cdot \kappa(M_{u|H})^4 \cdot \lambda_{\max}(w)/\beta(w) \Bigr)
\cdot \|\hC_l -\tilC_l\|_2
\nonumber.
\end{align}
By Lemma~\ref{lemma:matrix-perturb} (as applied to $\hM^n_{u,w,\{S;k\}}$
and $M_{u,w,\{S;k\}}$), we have
\begin{align}\nn
\|\hM_{w|H,\{S;k\}} \bfe_j - M_{w|H,\{S;k\}} \bfe_{\tau(j)}\|_2
&\leq \|Z^{-1}\h\nubf^{(j)} -\hU_{w|H,\{S;k\}}^\t M_{w|H,\{S;k\}}\e_{\tau(j)}\|_2
\\  &+ 2\| M_{w|H,\{S;k\}} \bfe_{\tau(j)}\|_2 \cdot \frac{\|\hM^n_{u,w,\{S;k\}} -
M_{u,w,\{S;k\}}\|_2}{\sigma_r(M_{u,w,\{S;k\}})}
.
\end{align}
\eprf


\subsection{Analysis of $\FindComponents$}\label{proof:findcomp}

We now provide results for Procedure $\FindComponents$ by using the previous result, where $w$ is set to each node pair $a,b\in V\setminus\{u_*\}$. We  condition on the event that $\hG_{\cup}= G_{\cup}$, where $G_{\cup}:=\cup_{h\in [r]} G_h$ is the union of the component graph.

We now give concentration bounds for $\beta$ and $\lambda_{\max}$ in \eqref{eqn:beta} and \eqref{eqn:lambdamax}.
Define \begin{align} \alpha_{\min} &:=\min_{a,b\in V\setminus\{u_*\}}\min_{\substack{k\in \Yc^{|S|}, |S|\leq 2 \eta\\ S\subset V\setminus \{a,b,u_*\}}}\min_{i\neq i'}\norm{ M_{(a,b)|H,\{S;k\}}(\bfe_i-\bfe_{i'})}_2\label{eqn:upsilonmin}\\ \label{eqn:upsilonmax} \alpha_{\max}&:= \max_{a,b\in V\setminus\{u_*\}}\max_{\substack{k\in \Yc^{|S|}, |S|\leq 2 \eta\\ S\subset V\setminus \{a,b,u_*\}}}\max_{j\in [r]}\|M_{(a,b)|H,\{S;k\}}\bfe_j\|_2, \end{align} and let \beq \alpha:=\frac{\alpha_{\max}}{\alpha_{\min}}.\label{eqn:alpha}\eeq

\begin{lemma}[Bounds for $\beta$ and $\lambda_{\max}$]Fix $\delta\in (0,1)$, given any $a,b\in V\setminus \{u_*\}$ and any set $S\subset V\setminus \{a,b,u_*\}$ with $|S|\leq 2\eta$, we have with probability at least $1-\delta$,
\begin{align}\beta(a,b) &\geq \frac{  \alpha_{\min} \,. \delta}{2\sqrt{er} {r \choose
2} r p^2 (pd)^{2\eta}}\\ \lambda_{\max}(a,b)& \leq \frac{\alpha_{\max}}{\sqrt{r}}
\Bigl(1 + \sqrt{2\ln(r^2 p^2 (pd)^{2\eta} /\delta)} \Bigr)
\end{align} This implies that with probability at least $1-2\delta$, \beq \frac{\lambda_{\max}(a,b)}{\beta(a,b)} \geq \frac{\sqrt{e}\alpha}{\delta} r^3 p^2 (pd)^{2\eta}\Bigl(1 + \sqrt{2\ln(r^2 p^2 (pd)^{2\eta} /\delta)} \Bigr),\eeq where $\alpha$ is given by \eqref{eqn:alpha}.
\end{lemma}

Similarly, we have bounds on $\norm{\hM^n_{u_*, a,b, \{S;k\}}- M_{u_*, a,b, \{S;k\}}}_2$ using Lemma~\ref{lemma:spectconc} and union bound.

\begin{proposition}[$\norm{\hM^n_{u_*, a,b, \{S;k\}}- M_{u_*, a,b, \{S;k\}}}_2$]With probability at least $1-\delta$, we have, for all $a, b\in V\setminus \{u_*\}$, $S\subset V\setminus \{a,b,u_*\}$, $|S|\leq 2\eta$, \beq \norm{\hM^n_{u_*, a,b, \{S;k\}}- M_{u_*, a,b, \{S;k\}}}_2 \leq \frac{1}{\sqrt{n}} \left(1+\sqrt{\log\left(\frac{p^{2\eta+2}d^{2\eta}}{\delta}\right)
}\right).\eeq\end{proposition}

Define $\rho'_{1,\min}$, $\rho'_{2, \min}$ and $\rho'_{\max}$ as
\begin{align}\rho'_{1,\min}&:=\min_{\substack{S\subset V\setminus\{u_*,v\}\\ |S|\leq 2\eta, k\in \Yc^{|S|}}}\min_{v\in V\setminus\{u_*\}} \sigma_r\left(M_{u_*, v, \{S;k\}}\right),\\ \rho'_{2,\min}&:=\min_{\substack{S\subset V\setminus\{u_*,a,b\}\\ |S|\leq 2\eta, k\in \Yc^{|S|}}}\min_{a,b\in V\setminus\{u_*\}} \sigma_r\left(M_{u_*, a,b, \{S;k\}}\right).\end{align}
Using the above defined constants, define
\begin{align}\nn K'(\delta;p,d,r):= &
1024  \cdot \kappa(M_{u|H})^4 \cdot \frac{\sqrt{e}\alpha}{\delta \rho'_{1,\min}} r^5 p^2 (pd)^{2\eta}\left(1 + \sqrt{2\ln(r^2 p^2 (pd)^{2\eta} /\delta)} \right) \\ &+48\frac{\sqrt{r}}{\rho'_{1,\min}} \cdot \kappa(M_{u|H})^2 + \frac{2\alpha_{\max}}{\rho'_{2,\min}},\label{eqn:K'}\end{align} and \beq \label{eqn:K} K(\delta;p,d,r):= K'(\delta;p,d,r) \left(1+\sqrt{\log\left(\frac{p^{2\eta+2}d^{2\eta}}{\delta}\right)
}\right).\eeq
We can now provide the final bound on distortion of estimated statistics using all the previous results.


\begin{lemma}[Bounds for $\|\hM_{a,b|H,\{S;k\}} \bfe_j - M_{a,b|H,\{S;k\}} \bfe_{\tau(j)}\|_2$]For any $a, b\in V\setminus \{u_*\}$, $k \in \Yc^{|S|}$, $j \in [r]$, there exists a permutation $\tau(j)\in [r]$ such that, conditioned on event that $\hG_{\cup}=G_{\cup}$, with probability at least $1-3\delta$, \beq\|\hM_{a,b|H,\{S;k\}} \bfe_j - M_{a,b|H,\{S;k\}} \bfe_{\tau(j)}\|_2\leq  \frac{K(\delta;p,d,r)}{\sqrt{n}} .\eeq This implies\beq\|\hM_{a,b |H} \bfe_j - M_{a,b|H} \bfe_{\tau(j)}\|_2\leq  \frac{K(\delta;p,d,r)}{\sqrt{n}} + \frac{K(\delta;p,d,r)}{K'(\delta;p,d,r)\sqrt{n}} \leq \frac{2K(\delta;p,d,r)}{\sqrt{n}}. \eeq\end{lemma}



\paragraph{Results on Random Rotation Matrix: }
We also require the following result  from~\citep{AnandkumarHsuKakade:multiview12}. The standard inner product between vectors $\v{u}$ and $\v{v}$ is denoted
by $\dotp{\v{u},\v{v}} = \v{u}^\t \v{v}$.
Let $\sigma_i(A)$ denote the $i^{\tha}$ largest singular value of a matrix $A$.
Let $\sphere^{m-1} := \{ \v{u} \in \R^m : \|\v{u}\|_2 = 1 \}$
denote the unit sphere in $\R^m$.
Let $\e_i \in \R^d$ denote the $i^{\tha}$ coordinate vector where the $i^{\tha}$
entry is $1$, and the rest are zero.

\begin{lemma} \label{lemma:eigen-gap}
Fix any $\delta \in (0,1)$ and matrix $A \in \R^{m \times n}$ (with $m \leq
n$).
Let $\v\theta \in \R^m$ be a random vector distributed uniformly over
$\sphere^{m-1}$.
\begin{enumerate}
\item $\displaystyle\Pr\biggl[ \min_{i \neq j} |\dotp{\v\theta,A(\e_i -
\e_j)}| > \frac{\sqrt2 \sigma_m(A) \cdot \delta}{\sqrt{em} {n \choose 2}}
\biggr] \geq 1-\delta$.

\item $\displaystyle\Pr\biggl[ \forall i \in [m],\ |\dotp{\v\theta,A\e_i}|
\leq \frac{\|A\e_i\|_2}{\sqrt{m}} \Bigl(1 + \sqrt{2\ln(m/\delta)} \Bigr)
\biggr] \geq 1-\delta$.

\end{enumerate}
\end{lemma}

\subsection{Improved Results for Tree Mixtures}\label{sec:treemixture}

We now consider a simplified version of $\FindComponents$ by limiting to estimation of pairwise marginals only on the edges of $\hG_{\cup}$, where $\hG_{\cup}$ is the estimate of $G_{\cup}:=\cup_{h\in [r]} G_h$, which is the union of the component graph, as well as constructing the Chow-Liu trees $\hT_h$ as subgraphs of $\hG_{\cup}$. Thus, instead of considering each node pair $a,b\in V\setminus \{u_*\}$, we only need to choose $(a,b)\in \hG_{\cup}$. Moreover, instead of considering $S\subset V\setminus \{a,b,u_*\}$, we can follow the convention of choosing $S\subset \nbd(a;\hG_{\cup})\cup \nbd(b;\hG_{\cup})$, and this changes the definition of $\alpha_{\min}$, $\alpha_{\max}$, $\rho'_{1, \min}$, $\rho'_{2,\min}$ and so on.   For all $(a,b)\in G_{\cup}$,  let \beq \Delta_2:=\max_{(a,b)\in G_{\cup}} |\nbd(a;G_{\cup})\cup \nbd(b;G_{\cup})|.\eeq We have improved bounds for $\beta$ and $\lambda_{\max}$ defined in \eqref{eqn:beta} and \eqref{eqn:lambdamax}, when $\Delta_2$ is small.

\begin{lemma}[Improved Bounds for $\beta$ and $\lambda_{\max}$]Fix $\delta\in (0,1)$,  when  $|S|\leq 2\eta$ and $S\subset \nbd(a;G_{\cup})\cup \nbd(b;G_{\cup})$, with probability at least $1-\delta$,
\begin{align}\beta(w) &\geq \frac{\sqrt2 \alpha_{\min}  \delta}{\sqrt{er} {r \choose
2} r p^2 d^{2\eta} \Delta_2^{2\eta}}\\ \lambda_{\max}(w)& \leq \frac{\alpha_{\max}}{\sqrt{r}}
\Bigl(1 + \sqrt{2\ln(r^2 p^2 d^{2\eta}\Delta_2^{2\eta} /\delta)} \Bigr)
\end{align}
\end{lemma}

We can substitute the above result to obtain a better bound $K^{\tree}(\delta;p, d, r)$ for learning tree mixtures.

\subsection{Analysis of Tree Approximations: Proof of Theorem~\ref{thm:treeapprox}}\label{sec:MI}

We now relate the perturbation of probability vector to perturbation of the corresponding mutual information~\citep{Cover&Thomas:book}. Recall that for discrete random variables $X,Y$,  the mutual information $I(X;Y)$ is related to their entropies $H(X,Y)$, $H(X)$ and $H(Y)$ as \beq I(X;Y) = H(X) + H(Y) - H(X,Y),\eeq and the entropy is defined as \beq H(X):= - \sum_{x\in \Xc} P(X=x)\log P(X=x),\eeq where $\Xc$ is the sample space of $X$. We recall the following result from~\citep{Shamiretal}.
Define function $\phi(x)$ for $ x\in \R^+$ as
\bsbcase{\phi(x)=\label{eqn:phi}}0, & $x=0$,\\ -x \log x, & $x \in (0, 1/e)$, \\ 1/e, & o.w.   \esbcase
\begin{proposition}For any $a,b\in [0,1]$, \beq |a\log a - b\log b| \leq \phi(|a-b|),\eeq for $\phi(\cdot)$ defined in \eqref{eqn:phi}.
\end{proposition}

We can thus prove bounds on  the estimated mutual information $\hI^{\spect}(\cdot)$ using statistics $\hP^{\spect}(\cdot)$ obtained from spectral decomposition.

\begin{proposition}[Bounding $|\hI^{\spect}(\cdot)- I(\cdot)|$] Under the event that $\|\hP^{\spect}(Y_a, Y_a|H=h)- P(Y_a, Y_a|H=h)\|_2 \leq \epsilon$, we have that
\beq |\hI^{\spect}(Y_a; Y_a|H=h)- I(Y_a; Y_a|H=h)| \leq 3 d \phi(\epsilon). \eeq\end{proposition}

For success of Chow-Liu algorithm, it is easy to see that the algorithm finds the correct tree when the estimated mutual information quantities are within half the minimum separation $\vartheta$ defined in \eqref{eqn:vartheta}. This is because the only wrong edges in the estimated tree $\hT_h$ are those that replace a certain edge in the original tree $T_h$, without violating the tree constraint. Similar ideas have been used by~\citet{Tan&etal:09ITsub} for deriving error exponent bounds for the Chow-Liu algorithm.
Define\beq \label{eqn:epsilontree} \epsilon^{\tree}:=\phi^{-1}\left(\frac{ 0.5 \vartheta -\tau}{3 d}\right).\eeq Thus, using the above result and assumption (A11) implies that we can estimate the mutual information to required accuracy to obtain the correct tree approximations.

\section{Analysis Under Local Separation Criterion}

%
%
%

\subsection{Rank Tests Under Approximate Separation}\label{proof:localrank-exactstat}

We now extend the results of the previous section when approximate separators are employed in contrast to exact vertex separators. Let $S:=\Sc_{\local}(u,v;G, \gamma)$ denote a local vertex separator  between any non-neighboring nodes $u$ and $v$ in graph $G$ under threshold $\gamma$.
We   note the following result on the  probability matrix $M_{u,v,\{S;k\}}$ defined in \eqref{eqn:M}.

\begin{lemma}[Rank Upon Approximate Separation]\label{lemma:localsep}Given a $r$-mixture of graphical models with $G=\cup_{k=1}^r G_k$, for any nodes $u,v\in V$ such that $\nbd[u]\cap \nbd[v]=\emptyset$ and $S:=\Sc_{\local}(u,v;G,\gamma)$ be any separator of $u$ and $v$ on $G$, the probability matrix $M_{u,v,\{S;k\}}:=[P[Y_u=i, Y_v=j,\bfY_S=k]]_{i,j}$ has effective rank at most $r$ for any $k \in \Yc^{|S|}$\beq \rank\left(M_{u,v,\{S;k\}}\,;  \zeta(\gamma)\right)\leq r, \quad \forall\, k\in \Yc^{|S|}, (u,v)\notin G,\eeq where $\zeta(\gamma):= 2\sqrt{d}\max_{h\in [r]}   \zeta_{h}(\gamma)$,    and $\zeta_{ h}(\cdot)$ is the correlation decay rate function in \eqref{eqn:corrdecay} corresponding to the model $P(\bfy|H=h)$ and $\gamma$ is the path threshold for local vertex separators.\end{lemma}

\noindent{\em Notation: }For convenience, for any node $v\in V$, let  $P(Y_v|H=h):=P(Y_v|H=h;G_h)$ denote the original component model Markov on graph $G_h$, and let $P(Y_v)$ denote the corresponding marginal distribution of $Y_v$ in the mixture. Let $\cP^\gamma(Y_v|H=h):= P(Y_v|H=h;F_{\gamma,h})$ denote the component model Markov on the induced subgraph $F_{\gamma,h}:= G_h(B_\gamma(v))$, where $B_{\gamma}(v;G_h)$ is the $\gamma$-neighborhood of node $v$ in $G_h$. In other words, we limit the model parameters up to $\gamma$ neighborhood and remove rest of the edges to obtain $\cP^\gamma(Y_v|H=h)$.

\vspace{1em}

\bprf
We first claim that
\beq\label{eqn:claim}\norm{M_{u|v,\{S;k\}}- M_{u|H,\{S;k\}} M_{H|v, \{S;k\}}}_{2}\leq \zeta(\gamma).
 \eeq
Note the relationship between the joint and the conditional probability matrices: \beq \label{eqn:jointcond}M_{u,v,\{S;k\}}=M_{u|v,\{S;k\}}
\Diag(\pibf_{v, \{S;k\}}),\eeq where $\pibf_{v, \{S;k\}} := [P(Y_v=i, \bfY_S=k)]^\top_i$ is the probability vector and $\Diag(\cdot)$ is the diagonal matrix with the corresponding probability vector as the diagonal elements.
Assuming \eqref{eqn:claim} holds and applying \eqref{eqn:jointcond}, we have that \begin{align}\nn&\norm{M_{u, v,\{S;k\}}- M_{u|H,\{S;k\}}M_{H|v, \{S;k\}}\Diag(\pibf_{v, \{S;k\}})}_{2}\\&\leq  \norm{\Diag(\pibf_{v, \{S;k\}})}_{2}  \zeta (\gamma)\leq \zeta(\gamma) ,\label{eqn:claim2} \end{align} since $\norm{\Diag(\pibf_{v, \{S;k\};G})}_{2}\leq \norm{\Diag(\pibf_{v, \{S;k\};G})}_{\Fbb}=\norm{\pibf_{v, \{S;k\};G}}_2 \leq 1$ for a probability vector.
From Weyl's theorem, assuming that \eqref{eqn:claim2} holds, we have\[ \rank\left(M_{u,v,\{S;k\}}\,;\, \zeta(\gamma)\right)\leq \min(r, d)=r,\] since we assume $r<d$ (assumption (B1) in Section~\ref{sec:assumptions-corrdecay}). Note that $\rank(A;\xi)$ denotes the effective rank, i.e., the number of singular values of $A$ which are greater than $\xi\geq 0$.

We now prove the claim in \eqref{eqn:claim}.
Since $G=\cup_{h=1}^r G_h$, we have that the resulting set   $S:=\Sc_{\local}(u,v;G,\gamma)$ is also a local separator on each of the component subgraphs $\{G_h\}_{h\in [r]}$ of $G$, for all sets $A, B\subset V$ such that $\nbd[u;G]\cap \nbd[v;G]=\emptyset$. Thus, we have that for all $k \in \Yc^{|S|}$,  $y_v\in \Yc$, $h \in [r]$,
\beq  \cP^\gamma(Y_u |Y_v=y_v,\bfY_S=k,H=h)  = \cP^\gamma(Y_u |\bfY_S=k,H=h).\label{eqn:localsep}\eeq  The statement in \eqref{eqn:localsep} is due to the fact that the nodes $u$ and $v$ are exactly separated by set $S$ in the subgraph $F_{\gamma,h}(u)$.

By assumption (B4) on correlation decay we have that
\[\norm{P(Y_u| Y_v=y_v,\bfY_S=k,H=h)-\cP^\gamma(Y_u|Y_v=y_v,\bfY_S=k,H=h)}_1 \leq \zeta_{h}(\gamma),\]for all $y_v\in \Yc$, $k \in \Yc^{|S|}$ and $h\in [r]$. Similarly, we also have
\[\norm{P(Y_u| \bfY_S=k,H=h)-\cP^\gamma(Y_u| \bfY_S=k,H=h)}_1 \leq \zeta_{h}(\gamma),\] which implies that \[\norm{P(Y_u| Y_v=y_v,\bfY_S=k,H=h)-P(Y_u|\bfY_S=k,H=h)}_1 \leq 2\zeta_{h}(\gamma),\]for all $y_v\in \Yc$, $k \in \Yc^{|S|}$ and $h\in [r]$, and thus, \beq\label{eqn:norm1} \norm{ M_{u|v, \{S;k\}} - M_{u|H, \{S;k\}} M_{H|v, \{S;k\}} }_1 \leq 2\max_{h\in [r]} \zeta_h(\gamma), \eeq where $\norm{A}_1$ of a matrix is the maximum column-wise absolute sum. Since $\norm{A}_2\leq \sqrt{d}\norm{A}_1$,
\eqref{eqn:claim} follows.\eprf

\subsection{Spectral Decomposition Under Local Separation}

We now extend the above analysis of spectral decomposition when a local separator is used instead of approximate separators. For simplicity consider nodes $u_*, a,b,c\in V$ (the same results can also be proven for larger sets), where $u_*$ is an isolated node in $G_{\cup}$, $a,b\in V\setminus \{u_*\}$, $c \notin \nbd[a;G_{\cup}]\cup \nbd[b;G_{\cup}]$ and let $S:= \Sc_{\local}((a,b), c;G_{\cup})$ be a local separator in $G_{\cup}$ separating $a,b$ from $c$. Since   we have \[ Y_{u_*}\indep \bfY_{V\setminus \{u_*\}} |H,\] the following decomposition holds
\[ M_{u_*,c,\{S;k\}} = M_{u_*|H} \Diag(\pibf_{H, \{S;k\}})M^\top_{c|H, \{S;k\}}.\]However, the matrix $M_{u_*, c,\{S;k\}, \{(a,b);q\}}$ no longer has a similar decomposition. Instead define
\beq \tilM_{u_*, c,\{S;k\}, \{(a,b);q\}}:= M_{u_*|H} \Diag(\pibf_{H, \{S;k\}, \{(a,b);q\}})M^\top_{c|H, \{S;k\}}.\eeq Define the observable operator, on lines of \eqref{eqn:tilC1}, based on $\tilM$ above rather than the actual probability matrix $M$, as \beq \label{eqn:tilC2}\widetilde{\tilC}(\bfm):= \left(U_1^\t \left(\sum_{q}m (q) \tilM_{u_*,c,\{S;k\},\{(a,b);q\}}\right) U_2\right) \left(U_1^\t M_{u_*, c, \{S;k\}}
U_2\right)^{-1},\eeq where $U_1$ is a matrix such that $U_1^\top M_{u_*|H}$ is invertible and $U_2$ is such that $U_2^\top M_{v|H, \{S;k\}}$ is invertible. On lines of Lemma~\ref{lemma:decompexact}, we have that \beq \widetilde{\tilC}(\bfm)= \left(U_1^T M_{u_*|H}\right)\Diag\left(M^\top_{(a,b)|H,\{S;k\}} \bfm\right)\left(U_1^T M_{u_*|H}\right)^{-1}.\eeq
Thus, the $r$ roots of the polynomial $\lambda \mapsto
\det(\widetilde{\tilC}(\bfm) - \lambda I)$ are
$\{ \dotp{\bfm,M_{(a,b)|H, \{S;k\} }\bfe_j} : j \in [r] \}$.
We now have show that $M$ and $\tilM$ are close under correlation decay.


\begin{proposition}[Regime of Correlation Decay]For all $k \in \Yc^{|S|}$ and $q \in \Yc^2$, we have \beq \norm{\tilM_{u_*, c,\{S;k\}, \{(a,b);q\}}-  M_{u_*, c,\{S;k\}, \{(a,b);q\}}}_2\leq \zeta(\gamma),\eeq where $\zeta(\gamma)$ is given by \eqref{eqn:zetatot}. \end{proposition}

\bprf On lines of obtaining \eqref{eqn:norm1} in the  proof of Lemma~\ref{lemma:localsep}, it is easy to see that \[ \norm{ P(Y_c| \bfY_S=k, \bfY_{a,b}=q) -\sum_{h \in [r]} P(Y_c| \bfY_S=k, H=h) P(H=h| \bfY_S=k, \bfY_{a,b}=q) }_1 \leq 2\max_{h\in [r]} \zeta_h(\gamma). \] This implies that for all $y \in \Yc$,  \begin{align}\nn  \|&   \sum_{h \in [r]}P(Y_{u_*}=y|H=h) P(H=h, \bfY_S=k, \bfY_{a,b}=q) P(Y_c| \bfY_S=k, H=h)\\ &-  P(Y_c, Y_{u_*}=y, \bfY_S=k, \bfY_{a,b}=q)\|_1\leq 2\max_{h\in [r]} \zeta_h(\gamma).\end{align} This is the same as
 \beq \norm{\tilM_{u_*, c,\{S;k\}, \{(a,b);q\}}-  M_{u_*, c,\{S;k\}, \{(a,b);q\}}}_\infty \leq 2\max_{h \in [r]} \zeta_h(\gamma), \eeq where $\norm{A}_\infty$ is the maximum absolute row sum and $\norm{A}_2\leq \sqrt{d} \norm{A}_\infty$ for a $d\times d$ matrix, and thus, we have the result.\eprf

\subsection{Spectral Bounds under Local Separation}

The result follows on similar lines as Section~\ref{proof:findcomp}, except that the distortion between the sample version of the observable operator $\hC(\bfm)$ and the desired version $\widetilde{\tilC}(\bfm)$ changes.  This leads to a slightly different bound

\begin{lemma}[Bounds for $\|\hM_{a,b|H,\{S;k\}} \bfe_j - M_{a,b|H,\{S;k\}} \bfe_{\tau(j)}\|_2$]For any $a, b\in V\setminus \{u_*\}$, $k \in \Yc^{|S|}$, $j \in [r]$, there exists a permutation $\tau(j)\in [r]$ such that, conditioned on event that $\hG_{\cup}=G_{\cup}$, with probability at least $1-3\delta$, \beq\|\hM_{a,b|H,\{S;k\}} \bfe_j - M_{a,b|H,\{S;k\}} \bfe_{\tau(j)}\|_2\leq  \frac{K(\delta;p,d,r)}{\sqrt{n}}  + K'(\delta;p,d,r) \zeta(\gamma),\eeq where $K'$ and $K$ are given by \eqref{eqn:K'} and \eqref{eqn:K}, and $\zeta(\gamma)$ is given by \eqref{eqn:zetatot}. This implies\beq\|\hM_{a,b |H} \bfe_j - M_{a,b|H} \bfe_{\tau(j)}\|_2\leq     \frac{2K(\delta;p,d,r)}{\sqrt{n}} +2 K'(\delta;p,d,r) \zeta(\gamma).\eeq\end{lemma}

\section{Matrix perturbation analysis}

We borrow the following results on matrix perturbation bounds from~\citep{AnandkumarHsuKakade:multiview12}.
We denote the $p$-norm of a vector $\v{v}$ by $\|\v{v}\|_p$, and the
corresponding induced norm of a matrix $A$ by $\|A\|_p := \sup_{\v{v} \neq
\v0} \|A\v{v}\|_p / \|\v{v}\|_p$.
The Frobenius norm of a matrix $A$ is denoted by $\|A\|_\Fbb$. For a matrix $A \in \R^{m \times n}$, let $\kappa(A) := \sigma_1(A) /
\sigma_{\min(m,n)}(A)$ (thus $\kappa(A) = \|A\|_2 \cdot \|A^{-1}\|_2$ if $A$
is invertible).


\begin{lemma} \label{lemma:matrix-perturb}
Let $X \in \R^{m \times n}$ be a matrix of rank $k$.
Let $U \in \R^{m \times k}$ and $V \in \R^{n \times k}$ be matrices with
orthonormal columns such that $\range(U)$ and $\range(V)$ are spanned by,
respectively, the left and right singular vectors of $X$ corresponding to
its $k$ largest singular values.
Similarly define $\h{U} \in \R^{m \times k}$ and $\h{V} \in \R^{n \times
k}$ relative to a matrix $\h{X} \in \R^{m \times n}$.
Define $\eps_X := \|\h{X} - X\|_2$, $\veps_0 :=
\frac{\eps_X}{\sigma_k(X)}$, and $\veps_1 := \frac{\veps_0}{1-\veps_0}$.
Assume $\veps_0 < \frac{1}{2}$.
Then
\begin{enumerate}
\item $\veps_1 < 1$;

\item $\sigma_k(\h{X}) = \sigma_k(\h{U}^\t \h{X} \h{V}) \geq (1-\veps_0)
\cdot \sigma_k(X) > 0$;

\item $\sigma_k(\h{U}^\t U) \geq \sqrt{1-\veps_1^2}$;

\item $\sigma_k(\h{V}^\t V) \geq \sqrt{1-\veps_1^2}$;

\item $\sigma_k(\h{U}^\t X \h{V}) \geq (1-\veps_1^2) \cdot \sigma_k(X)$;

\item for any $\h\alpha \in \R^k$ and
$\v{v} \in \range(U)$,
$\|\h{U}\h\alpha - \v{v}\|_2^2 \leq \|\h\alpha - \h{U}^\t
\v{v}\|_2^2 + \|\v{v}\|_2^2 \cdot \veps_1^2$.

\end{enumerate}
\end{lemma}

\begin{lemma} \label{lemma:op-perturb}
Consider the setting and definitions from Lemma~\ref{lemma:matrix-perturb},
and let $Y \in \R^{m \times n}$ and $\h{Y} \in \R^{m \times n}$ be given.
Define $\veps_2 := \frac{\veps_0}{(1-\veps_1^2) \cdot (1- \veps_0 -
\veps_1^2)}$ and $\eps_Y := \|\h{Y} - Y\|_2$.
Assume $\veps_0 < \frac1{1+\sqrt2}$.
Then
\begin{enumerate}
\item $\h{U}^\t \h{X} \h{V}$ and $\h{U}^\t X \h{V}$ are both invertible,
and $\|(\h{U}^\t \h{X} \h{V})^{-1} - (\h{U}^\t X \h{V})^{-1}\|_2 \leq
\frac{\veps_2}{\sigma_k(X)}$;

\item $\|(\h{U}^\t \h{Y} \h{V}) (\h{U}^\t \h{X} \h{V})^{-1} - (\h{U}^\t Y
\h{V}) (\h{U}^\t X \h{V})^{-1}\|_2 \leq \frac{\eps_Y}{(1 - \veps_0) \cdot
\sigma_k(X)} + \frac{\|Y\|_2 \cdot \veps_2}{\sigma_k(X)}$.

\end{enumerate}

\end{lemma}

\begin{lemma} \label{lemma:eig-perturb}
Let $A \in \R^{k \times k}$ be a diagonalizable matrix with $k$ distinct
real eigenvalues $\lambda_1,\lambda_2,\dotsc,\lambda_k \in \R$
corresponding to the (right) eigenvectors $\v\xi_1, \v\xi_2, \dotsc,
\v\xi_k \in \R^k$ all normalized to have $\|\v\xi_i\|_2 = 1$.
Let $R \in \R^{k \times k}$ be the matrix whose $i^{\tha}$ column is $\v\xi_i$.
Let $\h{A} \in \R^{k \times k}$ be a matrix.
Define $\eps_A := \|\h{A} - A\|_2$,
$\gamma_A := \min_{i \neq j} |\lambda_i - \lambda_j|$,
and
$\veps_3 := \frac{\kappa(R) \cdot \eps_A}{\gamma_A}$.
Assume $\veps_3 < \frac12$.
Then there exists a permutation $\tau$ on $[k]$ such that the
following holds:
\begin{enumerate}
\item $\h{A}$ has $k$ distinct real eigenvalues $\h\lambda_1, \h\lambda_2,
\dotsc, \h\lambda_k \in \R$, and $|\h\lambda_{\tau(i)} - \lambda_i| \leq
\veps_3 \cdot \gamma_A$ for all $i \in [k]$;

\item $\h{A}$ has corresponding (right) eigenvectors $\h\xi_1, \h\xi_2,
\dotsc, \h\xi_k \in \R^k$, normalized to have $\|\h\xi_i\|_2 = 1$, which
satisfy $\|\h\xi_{\tau(i)} - \v\xi_i\|_2 \leq 4(k-1) \cdot \|R^{-1}\|_2
\cdot \veps_3$ for all $i \in [k]$;

\item the matrix $\h{R} \in \R^{k \times k}$ whose $i^{\tha}$ column is
$\h\xi_{\tau(i)}$ satisfies $\|\h{R}-R\|_2 \leq \|\h{R}-R\|_\Fbb \leq
4k^{1/2}(k-1) \cdot \|R^{-1}\|_2 \cdot \veps_3$.

\end{enumerate}

\end{lemma}

\begin{lemma} \label{lemma:eig-perturb-all}
Let $A_1, A_2, \dotsc, A_k \in \R^{k \times k}$ be diagonalizable matrices
that are diagonalized by the same matrix invertible $R \in \R^{k \times k}$
with unit length columns $\|R\e_j\|_2 = 1$, such that each $A_i$ has $k$
distinct real eigenvalues:
\[ R^{-1} A_i R = \diag(\lambda_{i,1}, \lambda_{i,2}, \dotsc,
\lambda_{i,k})
.
\]
Let $\h{A}_1, \h{A}_2, \dotsc, \h{A}_k \in \R^{k \times k}$ be given.
Define $\eps_A := \max_i \|\h{A}_i - A_i\|_2$, $\gamma_A := \min_i \min_{j
\neq j'} |\lambda_{i,j} - \lambda_{i,j'}|$,
$\lambda_{\max} := \max_{i,j} |\lambda_{i,j}|$,
$\veps_3 := \frac{\kappa(R) \cdot \eps_A}{\gamma_A}$,
and $\veps_4 := 4k^{1.5} \cdot \|R^{-1}\|_2^2 \cdot \veps_3$.
Assume $\veps_3 < \frac12$ and $\veps_4 < 1$.
Then there exists a permutation $\tau$ on $[k]$ such that the
following holds.
\begin{enumerate}
\item The matrix $\h{A}_1$ has $k$ distinct real eigenvalues
$\h\lambda_{1,1}, \h\lambda_{1,2}, \dotsc, \h\lambda_{1,k} \in \R$, and
$|\h\lambda_{1,j} - \lambda_{1,\tau(j)}| \leq \veps_3 \cdot \gamma_A$ for
all $j \in [k]$.

\item There exists a matrix $\h{R} \in \R^{k \times k}$ whose $j^{\tha}$ column
is a right eigenvector corresponding to $\h\lambda_{1,j}$, scaled so
$\|\h{R}\e_j\|_2 = 1$ for all $j \in [k]$, such that $\|\h{R} - R_\tau\|_2
\leq \frac{\veps_4}{\|R^{-1}\|_2}$, where $R_\tau$ is the matrix obtained
by permuting the columns of $R$ with $\tau$.

\item The matrix $\h{R}$ is invertible and its inverse satisfies
$\|\h{R}^{-1} - R_\tau^{-1}\|_2 \leq \|R^{-1}\|_2 \cdot
\frac{\veps_4}{1-\veps_4}$;

\item For all $i \in \{2,3,\dotsc,k\}$ and all $j \in [k]$,
the $(j,j)^{\tha}$ element of $\h{R}^{-1} \h{A}_i \h{R}$, denoted by
$\h\lambda_{i,j} := \e_j^\t \h{R}^{-1} \h{A}_i \h{R} \e_j$, satisfies
\begin{align*}
|\h\lambda_{i,j} - \lambda_{i,\tau(j)}|
& \leq
\biggl( 1 + \frac{\veps_4}{1-\veps_4} \biggr)
\cdot \biggl( 1 + \frac{\veps_4}{\sqrt{k} \cdot \kappa(R)} \biggr)
\cdot \veps_3 \cdot \gamma_A
\\
& \quad{}
+ \kappa(R)
\cdot \biggl( \frac{1}{1-\veps_4} + \frac{1}{\sqrt{k} \cdot \kappa(R)} +
\frac{1}{\sqrt{k}} \cdot \frac{\veps_4}{1-\veps_4} \biggr)
\cdot \veps_4 \cdot \lambda_{\max}
.
\end{align*}
If $\veps_4 \leq \frac12$, then $|\h\lambda_{i,j} - \lambda_{i,\tau(j)}|
\leq 3\veps_3 \cdot \gamma_A + 4\kappa(R) \cdot \veps_4 \cdot
\lambda_{\max}$.

\end{enumerate}
\end{lemma}

\begin{lemma} \label{lemma:normalize-eig}
Let $V \in \R^{k \times k}$ be an invertible matrix, and let $R \in \R^{k
\times k}$ be the matrix whose $j^{\tha}$ column is $V\e_j / \|V\e_j\|_2$.
Then $\|R\|_2 \leq \kappa(V)$, $\|R^{-1}\|_2 \leq \kappa(V)$, and
$\kappa(R) \leq \kappa(V)^2$.
\end{lemma}


\end{appendix}


\begin{thebibliography}{43}
\providecommand{\natexlab}[1]{#1}
\providecommand{\url}[1]{\texttt{#1}}
\expandafter\ifx\csname urlstyle\endcsname\relax
  \providecommand{\doi}[1]{doi: #1}\else
  \providecommand{\doi}{doi: \begingroup \urlstyle{rm}\Url}\fi

\bibitem[Anandkumar and Valluvan(2012)]{Anandkumar:girth12}
A.~Anandkumar and R.~Valluvan.
\newblock {Learning Loopy Graphical Models with Latent Variables: Efficient
  Methods and Guarantees}.
\newblock \emph{Preprint. Available on ArXiv:1203.3887}, Jan. 2012.

\bibitem[Anandkumar et~al.(2011)Anandkumar, Chaudhuri, Hsu, Kakade, Song, and
  Zhang]{AnandkumarEtal:spectral}
A.~Anandkumar, K.~Chaudhuri, D.~Hsu, S.M. Kakade, L.~Song, and T.~Zhang.
\newblock {Spectral Methods for Learning Multivariate Latent Tree Structure}.
\newblock \emph{Preprint, ArXiv 1107.1283}, July 2011.

\bibitem[Anandkumar et~al.(2012{\natexlab{a}})Anandkumar, Hsu, and
  Kakade]{AnandkumarHsuKakade:COLT12}
A.~Anandkumar, D.~Hsu, and S.M. Kakade.
\newblock {A Method of Moments for Mixture Models and Hidden Markov Models}.
\newblock In \emph{Proc. of Conf. on Learning Theory}, June 2012{\natexlab{a}}.

\bibitem[Anandkumar et~al.(2012{\natexlab{b}})Anandkumar, Hsu, and
  Kakade]{AnandkumarHsuKakade:multiview12}
A.~Anandkumar, D.~Hsu, and S.M. Kakade.
\newblock {A Method of Moments for Mixture Models and Hidden Markov Models}.
\newblock \emph{Preprint}, Feb. 2012{\natexlab{b}}.

\bibitem[Anandkumar et~al.(2012{\natexlab{c}})Anandkumar, Tan, Huang, and
  Willsky]{AnandkumarTanWillsky:Ising11}
A.~Anandkumar, V.~Y.~F. Tan, F.~Huang, and A.~S. Willsky.
\newblock {High-Dimensional Structure Learning of Ising Models: Local
  Separation Criterion}.
\newblock \emph{Accepted to Annals of Statistics}, Jan. 2012{\natexlab{c}}.

\bibitem[Armstrong et~al.(2009)Armstrong, Carter, Wong, and
  Kohn]{Armstrong:2009}
H.~Armstrong, C.~K. Carter, K.~F. Wong, and R.~Kohn.
\newblock Bayesian covariance matrix estimation using a mixture of decomposable
  graphical models.
\newblock \emph{Statistics and Computing}, 19:\penalty0 303--316, September
  2009.

\bibitem[Belkin and Sinha(2010)]{belkin2010polynomial}
M.~Belkin and K.~Sinha.
\newblock Polynomial learning of distribution families.
\newblock In \emph{IEEE Annual Symposium on Foundations of Computer Science},
  pages 103--112, 2010.

\bibitem[Br{\'e}maud(1999)]{Bremaud:book}
P.~Br{\'e}maud.
\newblock \emph{{Markov Chains: Gibbs fields, Monte Carlo simulation, and
  queues}}.
\newblock Springer, 1999.

\bibitem[Bresler et~al.(2008)Bresler, Mossel, and Sly]{Bresler&etal:Rand}
G.~Bresler, E.~Mossel, and A.~Sly.
\newblock {Reconstruction of Markov Random Fields from Samples: Some
  Observations and Algorithms}.
\newblock In \emph{Intl. workshop APPROX Approximation, Randomization and
  Combinatorial Optimization}, pages 343--356. Springer, 2008.

\bibitem[Chandrasekaran et~al.(2010)Chandrasekaran, Parrilo, and
  Willsky]{Chandrasekaran:10latent}
V.~Chandrasekaran, P.A. Parrilo, and A.S. Willsky.
\newblock {Latent Variable Graphical Model Selection via Convex Optimization}.
\newblock \emph{Preprint. Available on ArXiv}, 2010.

\bibitem[Chang(1996)]{chang1996full}
J.T. Chang.
\newblock Full reconstruction of markov models on evolutionary trees:
  identifiability and consistency.
\newblock \emph{Mathematical Biosciences}, 137\penalty0 (1):\penalty0 51--73,
  1996.

\bibitem[Chen et~al.(2008)Chen, Zhang, and Wang]{Che08}
T.~Chen, N.~L. Zhang, and Y.~Wang.
\newblock Efficient model evaluation in the search based approach to latent
  structure discovery.
\newblock In \emph{4th European Workshop on Probabilistic Graphical Models},
  2008.

\bibitem[Choi et~al.(2011)Choi, Tan, Anandkumar, and Willsky]{Choi&etal:10JMLR}
M.J. Choi, V.Y.F. Tan, A.~Anandkumar, and A.~Willsky.
\newblock {Learning Latent Tree Graphical Models}.
\newblock \emph{J. of Machine Learning Research}, 12:\penalty0 1771--1812, May
  2011.

\bibitem[Chow and Liu(1968)]{Chow&Liu:68IT}
C.~Chow and C.~Liu.
\newblock {Approximating Discrete Probability Distributions with Dependence
  Trees}.
\newblock \emph{IEEE Tran. on Information Theory}, 14\penalty0 (3):\penalty0
  462--467, 1968.

\bibitem[Cover and Thomas(2006)]{Cover&Thomas:book}
T.~Cover and J.~Thomas.
\newblock \emph{Elements of Information Theory}.
\newblock John Wiley \& Sons, Inc., 2006.

\bibitem[Dasgupta(1999)]{dasgupta1999learning}
S.~Dasgupta.
\newblock Learning mixtures of gaussians.
\newblock In \emph{Foundations of Computer Science, IEEE Annual Symposium on},
  1999.

\bibitem[Daskalakis et~al.(2006)Daskalakis, Mossel, and Roch]{daskalakis06}
C.~Daskalakis, E.~Mossel, and S.~Roch.
\newblock Optimal phylogenetic reconstruction.
\newblock In \emph{STOC '06: Proceedings of the thirty-eighth annual ACM
  symposium on Theory of computing}, pages 159--168, 2006.

\bibitem[Durbin et~al.(1999)Durbin, Eddy, Krogh, and Mitchison]{Durbin}
R.~Durbin, S.~R. Eddy, A.~Krogh, and G.~Mitchison.
\newblock \emph{Biological Sequence Analysis: Probabilistic Models of Proteins
  and Nucleic Acids}.
\newblock Cambridge Univ. Press, 1999.

\bibitem[Erd\"{o}s et~al.(1999)Erd\"{o}s, Sz\'{e}kely, Steel, and
  Warnow]{erdos99}
P.~L. Erd\"{o}s, L.~A. Sz\'{e}kely, M.~A. Steel, and T.~J. Warnow.
\newblock A few logs suffice to build (almost) all trees: Part i.
\newblock \emph{Random Structures and Algorithms}, 14:\penalty0 153--184, 1999.

\bibitem[Geiger and Heckerman(1996)]{geiger1996knowledge}
D.~Geiger and D.~Heckerman.
\newblock Knowledge representation and inference in similarity networks and
  bayesian multinets.
\newblock \emph{Artificial Intelligence}, 82\penalty0 (1-2):\penalty0 45--74,
  1996.

\bibitem[Guo et~al.(2011)Guo, Levina, Michailidis, and Zhu]{guo2011joint}
J.~Guo, E.~Levina, G.~Michailidis, and J.~Zhu.
\newblock Joint estimation of multiple graphical models.
\newblock \emph{Biometrika}, 98\penalty0 (1):\penalty0 1, 2011.

\bibitem[Hsu et~al.(2009)Hsu, Kakade, and Zhang]{hsu2008spectral}
D.~Hsu, S.M. Kakade, and T.~Zhang.
\newblock A spectral algorithm for learning hidden markov models.
\newblock In \emph{Proc. of COLT}, 2009.

\bibitem[Jalali et~al.(2011)Jalali, Johnson, and Ravikumar]{Jalali:greedy}
A.~Jalali, C.~Johnson, and P.~Ravikumar.
\newblock On learning discrete graphical models using greedy methods.
\newblock In \emph{Proc. of NIPS}, 2011.

\bibitem[Kumar and Koller(2009)]{kumar2009learning}
M.P. Kumar and D.~Koller.
\newblock Learning a small mixture of trees.
\newblock In \emph{Proc. of NIPS}, 2009.

\bibitem[Lauritzen(1996)]{Lauritzen:book}
S.~L. Lauritzen.
\newblock \emph{{Graphical models}}.
\newblock Clarendon Press, 1996.

\bibitem[Lazarsfeld and Henry(1968)]{lazarsfeld68}
P.~F. Lazarsfeld and N.W. Henry.
\newblock \emph{Latent structure analysis}.
\newblock Boston: Houghton Mifflin, 1968.

\bibitem[Lindsay(1995)]{lindsay1995mixture}
B.G. Lindsay.
\newblock Mixture models: theory, geometry and applications.
\newblock In \emph{NSF-CBMS Regional Conference Series in Probability and
  Statistics}. JSTOR, 1995.

\bibitem[Meila and Jordan(2001)]{meila2001learning}
M.~Meila and M.I. Jordan.
\newblock Learning with mixtures of trees.
\newblock \emph{J. of Machine Learning Research}, 1:\penalty0 1--48, 2001.

\bibitem[Meinshausen and B\"{u}hlmann(2006)]{Mei06}
N.~Meinshausen and P.~B\"{u}hlmann.
\newblock {High Dimensional Graphs and Variable Selection With the Lasso}.
\newblock \emph{Annals of Statistics}, 34\penalty0 (3):\penalty0 1436--1462,
  2006.

\bibitem[Moitra and Valiant(2010)]{moitra2010settling}
A.~Moitra and G.~Valiant.
\newblock Settling the polynomial learnability of mixtures of gaussians.
\newblock In \emph{IEEE Annual Symposium on Foundations of Computer Science},
  2010.

\bibitem[Mossel and Roch(2006)]{mossel2005learning}
E.~Mossel and S.~Roch.
\newblock {Learning nonsingular phylogenies and hidden Markov models}.
\newblock \emph{The Annals of Applied Probability}, 16\penalty0 (2):\penalty0
  583--614, 2006.

\bibitem[Mossel and Roch(2011)]{mossel2011phylogenetic}
E.~Mossel and S.~Roch.
\newblock Phylogenetic mixtures: Concentration of measure in the large-tree
  limit.
\newblock \emph{Arxiv preprint arXiv:1108.3112}, 2011.

\bibitem[Netrapalli et~al.(2010)Netrapalli, Banerjee, Sanghavi, and
  Shakkottai]{Sanghavi&etal:Allerton10}
P.~Netrapalli, S.~Banerjee, S.~Sanghavi, and S.~Shakkottai.
\newblock {Greedy Learning of Markov Network Structure }.
\newblock In \emph{Proc. of Allerton Conf. on Communication, Control and
  Computing}, Monticello, USA, Sept. 2010.

\bibitem[Ravikumar et~al.(2008)Ravikumar, Wainwright, and
  Lafferty]{Ravikumar&etal:08Stat}
P.~Ravikumar, M.J. Wainwright, and J.~Lafferty.
\newblock {High-dimensional Ising Model Selection Using l1-Regularized Logistic
  Regression}.
\newblock \emph{Annals of Statistics}, 2008.

\bibitem[Ravikumar et~al.(2011)Ravikumar, Wainwright, Raskutti, and
  Yu]{Ravikumar&etal:08Arxiv}
P.~Ravikumar, M.J. Wainwright, G.~Raskutti, and B.~Yu.
\newblock {High-dimensional covariance estimation by minimizing
  $\ell_1$-penalized log-determinant divergence}.
\newblock \emph{Electronic Journal of Statistics}, \penalty0 (4):\penalty0
  935--980, 2011.

\bibitem[Shamir et~al.(2008)Shamir, Sabato, and Tishby]{Shamiretal}
Ohad Shamir, Sivan Sabato, and Naftali Tishby.
\newblock Learning and generalization with the information bottleneck.
\newblock In \emph{Algorithmic Learning Theory}, volume 5254 of \emph{Lecture
  Notes in Computer Science}, pages 92--107. 2008.

\bibitem[Spirtes and Meek(1995)]{spirtes1995learning}
P.~Spirtes and C.~Meek.
\newblock Learning bayesian networks with discrete variables from data.
\newblock In \emph{Proc. of Intl. Conf. on Knowledge Discovery and Data
  Mining}, pages 294--299, 1995.

\bibitem[Tan et~al.(2011)Tan, Anandkumar, and Willsky]{Tan&etal:09ITsub}
V.Y.F. Tan, A.~Anandkumar, and A.~Willsky.
\newblock {A Large-Deviation Analysis for the Maximum Likelihood Learning of
  Tree Structures}.
\newblock \emph{IEEE Tran. on Information Theory}, 57\penalty0 (3):\penalty0
  1714--1735, March 2011.

\bibitem[Thiesson et~al.(1999)Thiesson, Meek, Chickering, and
  Heckerman]{thiesson1999computationally}
B.~Thiesson, C.~Meek, D.~Chickering, and D.~Heckerman.
\newblock Computationally efficient methods for selecting among mixtures of
  graphical models.
\newblock \emph{Bayesian Statistics}, 6:\penalty0 569--576, 1999.

\bibitem[Wainwright and Jordan(2008)]{Wainwright&Jordan:08NOW}
M.J. Wainwright and M.I. Jordan.
\newblock {Graphical Models, Exponential Families, and Variational Inference}.
\newblock \emph{Foundations and Trends in Machine Learning}, 1\penalty0
  (1-2):\penalty0 1--305, 2008.

\bibitem[Weitz(2006)]{Weitz:STOC06}
D.~Weitz.
\newblock {Counting independent sets up to the tree threshold}.
\newblock In \emph{Proc. of ACM symp. on Theory of computing}, pages 140--149,
  2006.

\bibitem[Zhang(2004)]{ZhangJMLR04}
N.~L. Zhang.
\newblock {Hierarchical Latent Class Models for Cluster Analysis}.
\newblock \emph{Journal of Machine Learning Research}, 5:\penalty0 697--723,
  2004.

\bibitem[Zhang and Kocka(2004)]{zhang04}
N.~L. Zhang and T~Kocka.
\newblock Efficient learning of hierarchical latent class models.
\newblock In \emph{ICTAI}, 2004.

\end{thebibliography}

\end{document}